%% file: main.tex
\documentclass[10pt,twocolumn,letterpaper]{article}
\usepackage[T1]{fontenc}
\usepackage{newtxtext,newtxmath}
\usepackage[margin=0.75in,columnsep=0.25in]{geometry}

\usepackage{algorithmic}
\usepackage{algorithm}
\usepackage{textcomp}
\usepackage{stfloats}
\usepackage{url}
\usepackage{verbatim}
\usepackage{cite}
\usepackage[table]{xcolor}
\usepackage[hidelinks]{hyperref}
\usepackage{bookmark}
\usepackage{graphicx}
\usepackage{booktabs}
\usepackage{longtable}
\usepackage{subcaption}
\usepackage{multirow}
\usepackage{amsmath}
 
\usepackage{array}
\usepackage{makecell}
\definecolor{sigOneColor}{RGB}{235,244,251}
\definecolor{sigTwoColor}{RGB}{198,222,239}
\definecolor{sigThreeColor}{RGB}{139,188,219}
\definecolor{sigNSColor}{RGB}{220,220,220}
\newcommand{\sigone}[1]{\cellcolor{sigOneColor}#1}
\newcommand{\sigtwo}[1]{\cellcolor{sigTwoColor}#1}
\newcommand{\sigthree}[1]{\cellcolor{sigThreeColor}#1}

\hypersetup{
  pdftitle={Silence and Noise: Resolving Suppression and Overvaluation in Multimodal Entity Alignment},
  pdfauthor={Xizhe Zhang, Jingfeng Zhang, Meng-Fen Chiang},
  pdfsubject={Multimodal entity alignment},
  pdfkeywords={multimodal entity alignment, knowledge graphs, ALMEA}
}

\begin{document}

\title{Silence and Noise: Resolving Suppression and Overvaluation in Multimodal Entity Alignment}

\author{Xizhe Zhang\textsuperscript{1}\quad
Jingfeng Zhang\textsuperscript{2}\quad
Meng-Fen Chiang\textsuperscript{3,*}}
\date{}

\twocolumn[{\begin{@twocolumnfalse}
\maketitle
\vspace{-1.8em}
\begin{center}\small
\textsuperscript{1}School of Computer Science, The University of Auckland, New Zealand\\
\textsuperscript{2}College of Intelligent Robotics and Advanced Manufacturing/\\
Institute of Trustworthy Embodied AI, Fudan University, China\\
\textsuperscript{3}College of Electrical and Computer Engineering, National Yang Ming Chiao Tung University, Taiwan\\[0.5em]
\href{mailto:xzha593@aucklanduni.ac.nz}{xzha593@aucklanduni.ac.nz}\quad
\href{mailto:zjf@fudan.edu.cn}{zjf@fudan.edu.cn}\quad
\href{mailto:ankechiang@gmail.com}{ankechiang@gmail.com}\\[0.3em]
\textsuperscript{*}Corresponding author\quad\textit{Preprint}
\end{center}

\begin{abstract}
Multimodal entity alignment (MMEA) is fundamental to integrating heterogeneous knowledge graphs (KGs), yet existing paradigms remain critically vulnerable to incomplete modalities and cross-graph structural asymmetries. This paper exposes a fundamental oversight in current MMEA research: treating feature missingness and topological discrepancies as isolated preprocessing or denoising challenges, which overlooks their destructive coupling during multimodal fusion. We systematically formulate and evaluate two newly identified failure modes: 1) \emph{Detail Missing (DM)-induced overvaluation}, where the absence of local modality details causes models to overemphasize uninformative, blurred representations; and 2) \emph{Context Imbalance (CI)-induced suppression}, where topological discrepancies across KGs allow dense structural contexts to overwhelm weaker but highly discriminative modal signals, triggering high-level semantic drift. To break this cycle of degradation, we introduce \textbf{ALMEA}, a generative and calibration framework that establishes a bidirectional feedback loop between low-level feature reconstruction and high-level latent space alignment. Specifically, Cross-Modality Imputation (CMI) synthesizes missing modality details to mitigate overvaluation, while Latent Semantic Calibration (LSC) aligns cross-graph joint embeddings to suppress topology-driven semantic drift. Under sparse supervision, we develop Controlled Iterative Selection (CIS) to adaptively acquire representative and diverse pseudo-labeled pairs, optimizing latent calibration under constrained budgets. Extensive evaluations on standard benchmarks demonstrate that ALMEA consistently outperforms state-of-the-art baselines. At a $20\%$ supervision rate, ALMEA yields substantial absolute improvements of up to $5.20\%$ in MRR and $6.40\%$ in Hits@1, maintaining robust gains across higher supervision thresholds. Our code is publicly available at \href{https://github.com/RTX4090123/ALMEA}{github.com/RTX4090123/ALMEA}.
\end{abstract}

\par\vspace{1.4em}\end{@twocolumnfalse}}]

\input{01-intro}
\input{02-literature}
\input{03-preliminart}
\input{04-model}
\input{05-experiment}

\section{Conclusion}
We introduced ALMEA, a multimodal entity alignment framework designed to address the systematic limitations of Detail Missing (DM) and Context Imbalance (CI). By analyzing the cross-modal interaction in incomplete MMKGs, we identified two coupled failure modes: the overvaluation of incomplete modalities and the topological suppression of low-density modalities. To mitigate these issues, ALMEA couples Cross-Modality Imputation (CMI) and Latent Semantic Calibration (LSC) within a bidirectional feedback loop. CMI reconstructs missing low-level details to counteract overvaluation, while LSC calibrates cross-graph joint distributions in the latent space to prevent semantic drift and suppress local contextual asymmetry. 
To ensure the viability of this calibration under sparse supervision, we integrated Controlled Iterative Selection (CIS), a mathematically rigorous iterative selection mechanism that optimizes candidate query pathways by balancing representativeness and spatial diversity under strict annotation budgets. Extensive experiments on standard benchmark datasets demonstrate that ALMEA consistently improves alignment accuracy and robustness compared to state-of-the-art baselines, demonstrating strong resilience to structural inconsistencies and severe modality missingness. In future work, we aim to investigate the generalizability of our latent calibration strategies to broader multimodal graph learning tasks and explore the integration of non-parametric density estimation to further reduce computational overhead.

\section*{Acknowledgments}
This work is partially supported by the National Science and Technology Council (NSTC), Taiwan (Grant NSTC-115-2221-E-A49-120-MY3).

\bibliographystyle{IEEEtran}
\bibliography{references}

\clearpage
\onecolumn
\appendix
\section*{Appendices}
\addcontentsline{toc}{section}{Appendices}
\input{06-Supplementary}

\end{document}

%% file: 01-intro.tex
\section{Introduction}\label{sec:intro}

\begin{figure*}[t]
    \centering

    \begin{subfigure}[t]{\textwidth}
        \centering
        \includegraphics[width=0.9\linewidth]{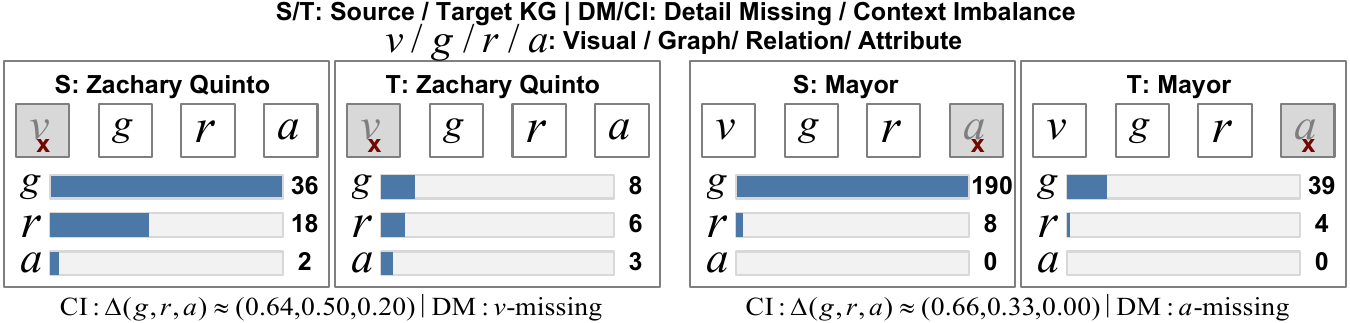}
        \label{fig:subfig-one-top}
    \end{subfigure}
    \vspace{4pt}
    
    \begin{subfigure}[t]{\textwidth}
        \centering
        \includegraphics[width=0.98\linewidth]{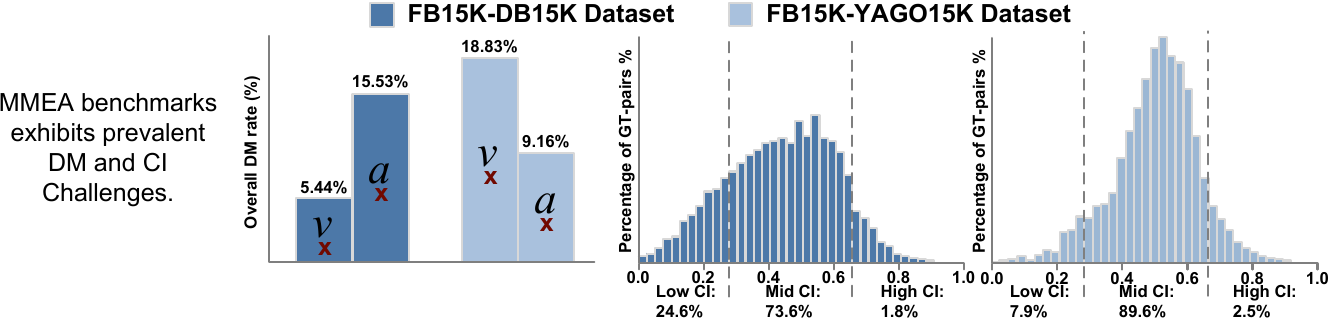}
        \label{fig:subfig-one-mid}
    \end{subfigure}

    \begin{subfigure}[t]{\textwidth}
        \centering
        \includegraphics[width=0.98\linewidth]{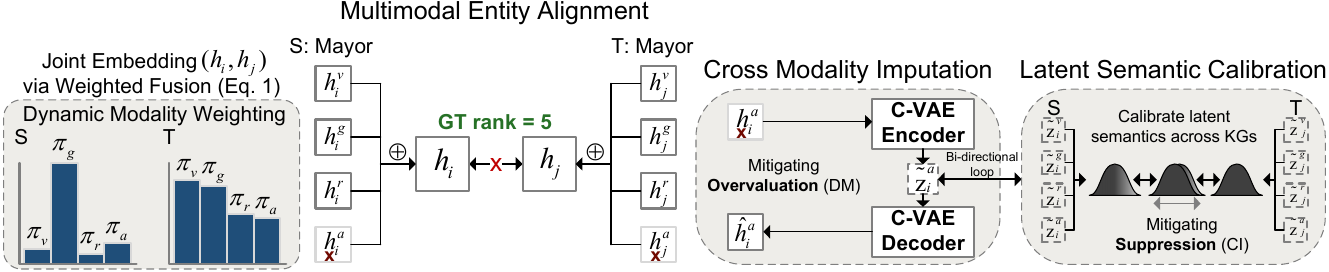}
        \label{fig:subfig-one-bottom}
    \end{subfigure}

    \caption{Overview of DM/CI challenges in MMEA benchmarks and the proposed ALMEA framework. \textbf{Top:} Two illustrative aligned entity pairs with Detail Missing (DM) and Context Imbalance (CI). ``x'' denotes a missing modality, and CI is measured using the modality-wise contextual discrepancies $\Delta(g,r,a)$ between ground-truth pairs across the source and target KGs. \textbf{Middle:} Dataset statistics on FB15K-DB15K and FB15K-YAGO15K: overall DM rates for visual/attribute modalities (left) and CI distributions over all ground-truth pairs (middle/right), grouped into Low/Mid/High regions by the averaged discrepancy score. \textbf{Bottom:} Our solutions to DM and CI. Entities are fused into joint embeddings $(h_i,h_j)$ with modality weights $\pi_m$ for nearest-neighbor retrieval. Cross-Modality Imputation (CMI) restores missing low-level details via a conditional VAE to mitigate DM-induced overvaluation, whereas Latent Semantic Calibration (LSC) aligns latent semantics across KGs to alleviate CI-induced suppression.}
    \label{fig:mainfig}
\end{figure*}

Multimodal Knowledge Graphs (MMKGs) have revolutionized modern knowledge representation by anchoring symbolic entities to multi-sensory contexts, integrating structural topology, textual attributes, and visual features. Consequently, Multimodal Entity Alignment (MMEA) has emerged as a cornerstone task for fusing these heterogeneous, distributed KGs into unified, coherent systems~\cite{chen2020mmea, liu2021visual, chen2022multi}. Traditional MMEA paradigms project relational structures, attribute descriptions, and visual modalities into a shared embedding space to compute entity similarities~\cite{chen2020mmea, lin2022multi}. Early frameworks leveraged Graph Convolutional Networks (GCNs) for cross-modal feature propagation~\cite{liu2021visual, chen2022multi}, while modern architectures employ contrastive learning~\cite{lin2022multi} and multi-head attention mechanisms~\cite{chen2023meaformer} to dynamically weight modality contributions. More recently, generative models such as GEEA~\cite{guo2023revisit} have introduced Variational Autoencoders (VAEs) to reconstruct missing features, seeking to heal incomplete data representations.

Despite these advancements, existing literature suffers from a fundamental oversight: it treats modality missingness as a localized reconstruction problem and views cross-graph topological variations as generic alignment noise~\cite{chen2023rethinking}. This compartmentalized view overlooks how these structural and modal deficiencies interact dynamically during multimodal fusion, leading to systemic representation degradation. In this work, we argue that incomplete MMKGs suffer from a dual-mode coupled failure mechanism, which we conceptualize as the interplay of \emph{Silence} (Detail Missing) and \emph{Noise} (Context Imbalance):

\noindent \textbf{DM-Induced Overvaluation (The Threat of Silence):} Detail Missing (DM) refers to the absence of entity-level modality details, such as missing images or attributes. Rather than merely presenting an incomplete input, the greater danger lies in \emph{overvaluing} uninformative placeholder representations. Standard models typically replace missing modalities with random vectors or global averages. During fusion, attention and weighting mechanisms often mistakenly assign significant importance to these blurred, semantically vacant vectors, biasing the joint representation and shifting the entity away from its true counterpart. As illustrated in Figure~\ref{fig:mainfig} (top), the missing visual information of ``Zachary Quinto'' and attribute information of ``Mayor'' exemplify this failure mode.

\noindent \textbf{CI-Induced Suppression (The Threat of Noise):} Context Imbalance (CI) represents the topological and structural discrepancy (disparities in neighborhood densities, relation distributions, and attribute diversity) between equivalent entities across KGs. During cross-modal fusion, highly dense contextual modalities generate over-represented, high-density embeddings. These dominant modalities completely suppress weaker, yet highly discriminative, low-density modalities (such as specific visual features or unique attributes), leading to severe semantic drift. For instance, Figure~\ref{fig:mainfig} shows that ``Mayor'' has 190 neighbors in the source KG but only 39 in the target KG, causing their contextual representations to carry substantially different information densities.

To resolve these coupled failure modes, we propose \textbf{ALMEA}, a generative MMEA framework that couples low-level modality reconstruction with high-level latent semantic calibration in a bidirectional feedback loop. ALMEA comprises three core components: 
(i) The \textbf{Cross-Modality Imputation (CMI)} module proactively simulates detail-missing scenarios during training using a conditional VAE. By learning to reconstruct missing low-level features from visible modalities, CMI mitigates the overvaluation of incomplete inputs.
(ii) The \textbf{Latent Semantic Calibration (LSC)} module aligns the cross-graph joint embeddings of counterpart entities, regulating the latent space to prevent neighborhood-driven semantic suppression. Crucially, CMI and LSC operate in a closed-loop: the restored modal details from CMI constrain the high-level semantic alignment, while the calibrated latent distributions from LSC guide subsequent feature reconstructions.
Finally, because high-level latent calibration requires rich supervision that is rarely available in real-world settings, we address resource constraints by refining self-bootstrapping iterative learning paradigms~\cite{sun2018bootstrapping}.
(iii) We propose \textbf{Controlled Iterative Selection (CIS)}, a mathematically formulated active acquisition rule. Unlike unconstrained bootstrapping methods that introduce cumulative pseudo-label noise, CIS selects highly representative, diverse, and low-CI candidate pairs by solving a combinatorial sparse subset selection objective via ADMM.
Extensive experiments validate ALMEA's effectiveness; we outperformed the strongest baselines by up to 5.20\% in MRR and 6.40\% in Hits@1 with 20\% supervision. The improvement remains significant under 50\% and 80\% supervision, with CIS making the greatest contribution under insufficient supervision. Ablation studies highlight its ability to bridge semantic gaps and handle incomplete MMKGs.

Our core contributions are summarized as follows:
\begin{itemize}
    \item We identify and define two coupled failure modes in MMEA under incomplete MMKGs. Detail Missing (DM)-induced overvaluation and Context Imbalance (CI)-induced suppression explain how low-level missingness and structural disparities amplify noise during fusion.
    \item We propose \textbf{ALMEA}, a bidirectional generative framework that dynamically couples Cross-Modality Imputation (CMI) and Latent Semantic Calibration (LSC) to simultaneously rectify local missing details and global semantic drift.
    \item We design a \textbf{Controlled Iterative Selection (CIS)} strategy using sparse optimization to systematically enrich the labeled training set with high-confidence, structurally diverse, and low-imbalance candidate pairs under restricted budgets.
    \item Extensive evaluations on standard benchmarks show that, under $20\%$ supervision, ALMEA yields significant improvements in MRR and Hits@1 by up to $5.20\%$ and $6.40\%$ over state-of-the-art baselines, demonstrating high robust gains across different low-resource and multilingual scenarios.
\end{itemize}

%% file: 02-literature.tex
\section{Related Work}\label{sec:work}
\noindent\textbf{Multimodal Entity Alignment. }
Entity alignment across KGs is fundamental for graph fusion and downstream reasoning. Structural GNNs (e.g., GCN, GAT) aggregate local topology and semantics to capture structural associations~\cite{kipf2017semi,velickovic2018graph}, while translation-based models (e.g., TransE, TransH) enforce triple-level geometric constraints in vector spaces~\cite{bordes2013translating,wang2014knowledge}. To incorporate richer data, Multimodal EA (MMEA) fuses cross-modal representations to score entity similarity~\cite{chen2023meaformer,chen2023rethinking,li2023multi,lin2022multi,chen2022multi,ni2023psnea,guo2023revisit,li2024simdiff,wang2025otmea}. Specifically, the approach in~\cite{chen2022multi} exploits visual semantics without GNNs, simplifying computation but weakening structural fidelity, whereas MEAformer~\cite{chen2023meaformer} couples GAT/GNN encoders with Transformer attention, which can become brittle under severe modality missingness. Similarly, SimDiff~\cite{li2024simdiff} maps multiple modalities into a unified latent space and applies Gaussian noise with progressive denoising to enable intermodal information transfer and reduce feature heterogeneity. Despite these developments, open challenges persist in improving robustness to modality drop/noise, establishing principled cross-modal fusion, performing cross-modal calibration, and enhancing sample efficiency.

\noindent\textbf{Missing Modality Imputation. }
Several studies~\cite{li2023attribute, guo2023revisit, chen2023rethinking} have explored modality imputation to handle incomplete MMKGs. For instance, the method proposed in~\cite{li2023attribute} leverages GNNs and generative algorithms for entity alignment; however, this framework is highly sensitive to input data quality and struggles to align latent cross-modal semantics. The generative framework in~\cite{guo2023revisit} applies a VAE for modality reconstruction but is constrained by source KG modality dependencies and limited semantic capture of the missing features. To mitigate these ambiguities, the paradigm in~\cite{chen2023rethinking} introduces an uncertainty-based method for approximating missing modalities, yet the overall reconstruction accuracy remains hindered by semantic ambiguity. 
However, these works focus primarily on reconstruction accuracy and do not address why modal embeddings with insufficient or ambiguous information are assigned excessive weights during multimodal fusion, leading to DM-induced overvaluation.

\noindent\textbf{Semantic Calibration.}
Key studies addressing semantic distribution discrepancies for cross-modal alignment include UNITER~\cite{chen2020uniter}, which uses a Transformer encoder for joint image-text embedding and masked semantic prediction. Similarly, ALBEF~\cite{li2021align} employs contrastive learning to align semantically similar pairs, whereas ALIGN~\cite{cohen1997align} maps features to probability distributions for distribution-level alignment. 
Nevertheless, existing calibration methods do not explicitly explain how cross-image contextual differences lead higher-information-density modalities to suppress weaker modalities, resulting in CI-induced suppression.
Building on these paradigms, we propose a latent semantic calibration module to bridge source-target semantic gaps and handle modality absence and inconsistency.

\noindent\textbf{From Iterative Learning to Active Learning.}
Recent entity alignment methods often rely on iterative learning to expand supervision from limited seed alignments. A representative example is BootEA~\cite{sun2018bootstrapping}, which augments training pairs with high-confidence pseudo-alignments predicted by the current model. While effective, this self-bootstrapping process inevitably introduces cumulative prediction noise. This limitation naturally motivates active learning, where the model selectively decides which high-confidence candidate pairs are worth querying for expert labeling.
Recent efforts have studied entity alignment in simulation-based active learning by selecting queries according to structural centrality, coverage, and uncertainty criteria~\cite{berrendorf2021active,liu2021activeea,zeng2021reinforced,huang2023deep}. For instance, RAC~\cite{zeng2021reinforced} uses a reinforcement learning policy over node degree, PageRank, and information density; the approach in~\cite{liu2021activeea} maximizes information gain via uncertainty scoring, and the method in~\cite{huang2023deep} refines candidate selection with similarity-margin constraints. However, these methods predominantly assume unimodal signals, lacking cross-modal semantic consistency and robustness to missing modalities. Moreover, their evaluations are typically confined to simulated, clean graph settings with stationary distributions, without addressing the necessary denoising or shift robustness, which limits their external validity in complex multimodal scenarios.

%% file: 03-preliminart.tex
\section{Preliminaries}\label{sec:background}
This section describes the multimodal entity alignment task and analyzes two key challenges encountered in real-world multimodal knowledge graphs (MMKGs): Detail Missing (DM) and Context Imbalance (CI).

\subsection{Problem Formulation}
\par\noindent \textbf {Multimodal Knowledge Graph (MMKG).}
A multimodal knowledge graph is formally represented as \( \mathcal{G} = \{\mathcal{E}, \mathcal{R}, \mathcal{A}, \mathcal{V}, \mathcal{T}\} \), where \( \mathcal{E} \) is the set of entities, \( \mathcal{R} \) denotes the set of relations, \( \mathcal{A} \) represents the set of attributes, and \( \mathcal{V} \) represents the visual information associated with the entities. The set \( \mathcal{T} \) consists of relational triples \( (e_h, r, e_t) \in \mathcal{T} \), where $e_h, e_t \in \mathcal{E}$ represent the head and tail entities, respectively, and $r \in \mathcal{R}$ denotes the relationship between them.

\par\noindent \textbf{Multimodal Entity Alignment (MMEA). }
Given a source MMKG, $\mathcal{G}_S = \{\mathcal{E}_S, \mathcal{R}_S, \mathcal{A}_S, \mathcal{V}_S, \mathcal{T}_S\}$, and a target MMKG, $\mathcal{G}_T = \{\mathcal{E}_T, \mathcal{R}_T, \mathcal{A}_T, \mathcal{V}_T, \mathcal{T}_T\}$, let $L = \{(e_s, e_t) \mid e_s \in \mathcal{E}_S, e_t \in \mathcal{E}_T, e_s \equiv e_t\}$ denote a seed set of known equivalent entity pairs, and let $U$ represent the set of entities with unknown counterparts. The goal of the MMEA task is to identify and align equivalent entity pairs between the unlabeled subsets of $\mathcal{G}_S$ and $\mathcal{G}_T$, which is formulated as identifying \( \{(e_i, e_j) \mid e_i \in \mathcal{E}^U_S, e_j \in \mathcal{E}^U_T, e_i \equiv e_j\} \), where \( \equiv \) denotes entity equivalence across MMKGs.

\par\noindent \textbf{Multimodal Knowledge Encoder. }
Each entity $e_i$ is typically represented by pre-trained feature embeddings $\{ x_i^v, x_i^g, x_i^r, x_i^a \}$ corresponding to the visual, graph neighbor, relational, and attribute modalities, respectively. Modality-specific encoders $\mathcal{M}^m$ project these input features to obtain modality embeddings $h_i^m = \mathcal{M}^m(x_i^m)$ for each modality $m \in \{v, g, r, a\}$~\cite{lin2022multi}. Following prior MMEA formulations (e.g.,~\cite{chen2023meaformer,li2024simdiff}), the cross-modal embedding $h_i$ of entity $e_i$ is obtained by projecting modality-specific features into a shared space and aggregating them via a normalized and weighted fusion:
\begin{equation}\label{eq:joint}
    h_i = \sum_{m \in \{v, g, r, a\}} \pi_m \cdot \text{Norm}(h^m_i), \pi_m = \frac{\exp(w_m)}{\sum_j \exp(w_j)},
\end{equation}
where the modality weight $\pi_m$ is dynamically derived from the learnable parameters $w_m$.

\subsection{Detail Missing (DM) and Overvaluation}\label{Detail Missing (DM)}
In practical MMKGs, complete modality profile availability is rarely guaranteed~\cite{wu2024deep}. To formally analyze this challenge, we introduce a modality availability indicator $\delta^m(e) \in \{0,1\}$, where $\delta^m(e)=0$ represents the absence of modality $m \in \{v, g, r, a\}$ for entity $e$, and $\delta^m(e)=1$ denotes its presence. To evaluate MMEA robustness under incomplete modalities, we define the overall Detail Missing (DM) rate of an entity set $\mathcal{E}$ across two standard benchmarks, FB15K-DB15K and FB15K-YAGO15K~\cite{Garcia-DuranN18}, as: \(\text{DM}(\mathcal{E})=\frac{1}{|\mathcal{E}||\mathcal{M}_{miss}|}\sum_{e\in\mathcal{E}}\sum_{m\in\mathcal{M}_{miss}} \mathbb{I}\!\left[\delta^{m}(e)=0\right],\)
where $\mathcal{E}\in\{\mathcal{E}_S,\mathcal{E}_T\}$ represents the source and target entity sets ($\mathcal{E}_S$ maps to FB15K, while $\mathcal{E}_T$ maps to DB15K or YAGO15K), and $\mathcal{M}_{miss} = \{v, a\}$ designates the subset of modalities susceptible to missingness, i.e., the visual and attribute modalities. Representative examples of visual and attribute missingness are depicted in Figure~\ref{fig:mainfig} (top), with their corresponding statistical distributions reported in Figure~\ref{fig:mainfig} (middle-left).

Empirical analysis indicates that DM occurs primarily within the visual and attribute modalities, $m \in \{v,a\}$. The absence of these modalities often leads to catastrophic alignment failures. Because these modalities contain rich entity-specific details, standard MMEA approaches~\cite{chen2023meaformer,liu2021visual} tend to bypass detailed entity-level imputation, opting instead to replace missing modalities with random vectors or global average embeddings. This fallback mechanism preserves significant noise and discards entity-level semantic details~\cite{yuan2010multiple}. Consequently, the fused joint embedding $h_i$ from Eq.~\ref{eq:joint} becomes semantically blurred, inducing severe alignment shifts. This systematic limitation motivates our Cross-Modality Imputation (CMI) module to reconstruct missing features prior to generating the joint entity embedding $h_i$, as shown in Figure~\ref{fig:mainfig} (bottom).

\subsection{Context Imbalance (CI) and Suppression}
Beyond the challenges introduced by DM, discrepancies in the local context between equivalent entities across KGs introduce severe semantic inconsistencies. CI describes the topological and structural variations in contextual information between ground-truth equivalent pairs across KGs. To evaluate the prevalence of CI in the FB15K-DB15K and FB15K-YAGO15K benchmarks, we define modality-wise discrepancies as $\Delta^{m}(e_i, e_j) = \frac{\left|c_{S}^{m}(e_i) - c_{T}^{m}(e_j)\right|} {c_{S}^{m}(e_i) + c_{T}^{m}(e_j) + \epsilon},$ where $\Delta^m$ represents the normalized contextual discrepancy of a ground-truth entity pair $(e_i, e_j)$ on modality $m \in \{g, r, a\}$. Specifically, $c^m(e)$ represents the context count: $c^g(e)$ denotes the number of unique neighbors (graph topology), $c^r(e)$ denotes the number of unique relation types, and $c^a(e)$ represents the count of distinct attributes. As shown in Figure~\ref{fig:mainfig} (top), these modality-wise differences $\Delta(g, r, a)$ directly capture structural asymmetry.

In Figure~\ref{fig:mainfig} (mid-mid, mid-right), we report the distribution of CI across all ground-truth aligned pairs (GT-pairs) in both benchmark datasets. The entity-pair level CI score is computed by averaging the modality-wise normalized discrepancies $\Delta^{m}(e_i, e_j)$ across $m \in \{g, r, a\}$:
\begin{equation}
    \text{CI}(e_i, e_j) = \frac{1}{3}\sum_{m \in \{g, r, a\}}\Delta^{m}(e_i, e_j)
\end{equation}
We categorize CI severity into three distinct intervals: Low CI ($\text{CI} \le 0.33$), Mid CI ($0.33 < \text{CI} \le 0.66$), and High CI ($\text{CI} > 0.66$), these thresholds are used to divide the confidence interval range into intervals of equal width. Significant topological imbalances are prevalent in both benchmarks: Mid and High CI GT-pairs constitute 75.4\% and 92.1\% of all equivalent pairs in FB15K-DB15K and FB15K-YAGO15K, respectively. This structural context imbalance across knowledge graphs motivates our Latent Semantic Calibration (LSC) module, designed to align cross-modal semantic representations and resolve CI-induced discrepancies, as shown in Figure~\ref{fig:mainfig} (bottom).

\subsection{Failure Analysis of DM and CI}\label{failure analysis subsection}
\begin{figure*}[t]
  \centering
  \includegraphics[width=\textwidth]{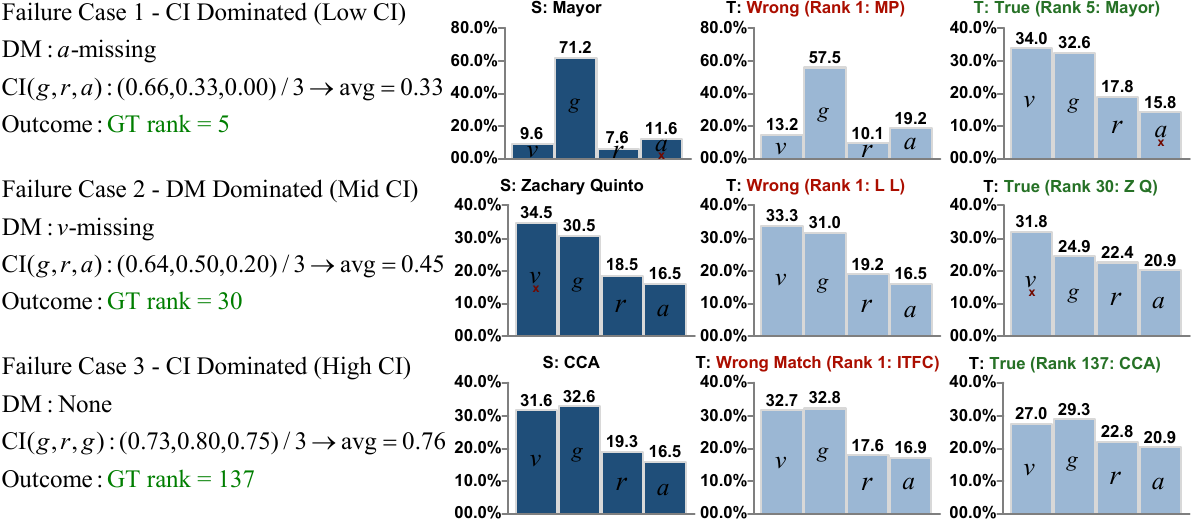}
  \caption{Representative failure cases of multimodal entity alignment under DM and CI. Each row shows modality weights $\pi_m$ ($m \in \{v,g,r,a\}$) for the source entity (\textbf{S}), the top-1 incorrect target match (\textbf{T}, wrong), and the ground-truth target at its retrieved rank (\textbf{T}, GT@r). Red ``x'' denotes a missing modality. Cases 1 and 3 are CI-dominated failures, whereas Case 2 is dominated by DM. Abbreviations: MP = "Member of Parliament", LL = "Lucy Liu", ZQ = "Zachary Quinto", CCA = "Chelsea College of Arts", and ITFC = "Ipswich Town F.C".}
  \label{fig:fig2_challenge}
\end{figure*}
The core objective of the MMEA task is to project entities into a shared space and identify equivalent counterparts based on nearest-neighbor search of their joint embeddings $h_i$~\cite{chen2020mmea,liu2021visual,chen2022multi, chen2023meaformer,vaswani2017attention,guo2023revisit}. Nearest-neighbor search algorithms are highly sensitive to high-weight modalities, meaning that dominant modalities can easily suppress contributions from weaker but highly discriminative ones~\cite{gong2025multimodal}. This causes the joint embedding $h_i$ to heavily favor modalities with excessively high weights $\pi_m$, directly distorting similarity rankings. To systematically investigate this behavior, we analyze the attention-based dynamic weighting system of MEAformer~\cite{chen2023meaformer}, which serves as a highly representative baseline for addressing DM and CI challenges.

As shown in Figure~\ref{fig:fig2_challenge}, we identify three distinct failure cases: 
(i) \textbf{Failure Case 1 (CI-Dominated):} Demonstrates the destructive impact of topological context imbalance. Although classified as Low CI overall, the graph neighbor discrepancy $\text{CI}(g)$ is highly severe (190 neighbors in the source KG versus only 39 in the target KG). This imbalance causes the graph modality weight $\pi_g$ for ``Mayor'' in the source KG to surge to 71.2\%, severely suppressing the visual, relational, and attribute modalities. The counterpart target entity, ``Member of Parliament'' (MP), suffers from a similar topological suppression, leading to significant semantic misalignment.
(ii) \textbf{Failure Case 2 (DM-Dominated):} Demonstrates the consequence of Detail Missing. Due to the absence of the visual modality ($v$-missing), standard mean or random imputation is applied. When constructing the joint embedding $h_i$ for ``Zachary Quinto'' (ZQ), the dynamic weighting system misinterprets this smoothed, uninformative visual feature as a reliable, noise-free information source~\cite{wu2024deep}. Consequently, $\pi_v$ is severely overestimated, causing the detail-lacking visual modality to be overvalued at the expense of other informative modalities.
(iii) \textbf{Failure Case 3 (High CI with Complete Modalities):} Even in the absence of missing modalities, severe topology and attribute discrepancies (High CI) degrade alignment accuracy, causing the ground-truth rank to drop to 137th—significantly worse than in the DM-dominated scenarios.

These systematic failures demonstrate that DM and CI significantly degrade alignment performance. To resolve these challenges, we propose the ALMEA framework (Figure~\ref{fig:mainfig}, bottom), which tightly couples CMI and LSC within a unified bidirectional loop. Leveraging a C-VAE, CMI first reconstructs the low-level details of missing modalities to prevent DM-induced overvaluation. Simultaneously, LSC calibrates the latent distribution of joint embeddings across source and target KGs, adjusting modality weights $\pi_m$ to suppress CI-driven semantic drift. Crucially, the calibrated high-level joint embeddings constrain low-level detail reconstruction, which in turn provides more precise signal inputs to correct high-level representations, forming a closed, mutually reinforcing loop.

\subsection{Mitigating CI under Limited Supervision}\label{Sec3:recoverd case analysis}
Because LSC relies on distribution matching in the latent space, its performance is highly dependent on abundant supervision. In low-resource scenarios, the benefits of latent semantic calibration can be severely limited by insufficient training data. To address this limitation, we improve upon conventional iterative self-training paradigms~\cite{sun2018bootstrapping} by proposing Controlled Iterative Selection (CIS). CIS actively identifies and labels the most representative and diverse candidate pairs within the current supervision environment under a highly restricted annotation budget, thereby maximizing the calibration efficiency of LSC. 
Further empirical analysis and qualitative recovery case studies under limited supervision are detailed in Section~\ref{supp:recovered_case_analysis} of the supplementary material.

%% file: 04-model.tex
\section{Methodology}\label{sec:4}
This section details the architectural formulation of ALMEA, designed to resolve the coupled dynamics of detail-missing overvaluation and context-driven suppression. By establishing a bidirectional, closed-loop learning paradigm, ALMEA couples low-level generative reconstruction with high-level latent semantic alignment, thereby decoupling these dual failure modes and preventing their mutual propagation.
As illustrated in the bottom panel of Figure~\ref{fig:mainfig}, ALMEA consists of three major components: Cross-Modality Imputation (CMI), Latent Semantic Calibration (LSC), and Controlled Iterative Selection (CIS). These modules are detailed in the following subsections.

\subsection{Cross-Modality Imputation (CMI)}\label{sec:4.1}
The overvaluation of incomplete modalities often stems from passive imputation paradigms~\cite{chen2023meaformer,lin2022multi} or uncalibrated generative reconstructions~\cite{guo2023revisit}. To address this limitation, the CMI module proactively simulates detail-missing scenarios during training via a Variational Autoencoder (VAE). By exposing the network to synthetic missing-modality perturbations of varying complexity, ALMEA learns modality embeddings that are highly robust to DM disturbances and can reconstruct missing low-level details.

\noindent\textbf{DM Disturbance Simulation. }
To formally simulate missing modality scenarios during training, we introduce a binary masking matrix $\mathbf{B} \in \{0,1\}^{d \times M}$, where $M$ denotes the number of modalities. Each column vector $b^m \in \{0,1\}^d$ represents a masking vector for modality $m \in \{v, g, r, a\}$, defining element-wise preservation or masking:
\begin{equation}\label{eq:masking}
    \tilde{h}^m = b^m \cdot h^m,\quad 
    b^m = \begin{cases} 
              1 & \text{if modality $m$ is preserved},\\
              0 & \text{if modality $m$ is masked}.
          \end{cases}
\end{equation}
where $\odot$ denotes the element-wise Hadamard product.

\noindent\textbf{Latent Embedding Construction. }
The masked modality embedding $\tilde{h}^m$ is fed into a modality-specific encoder and mapped to the mean and standard deviation of a Gaussian distribution. Using the reparameterization trick~\cite{kingma2013auto} to allow end-to-end backpropagation, the latent embedding $\tilde{z}^m$ is sampled from the parameterized Gaussian distribution as:
\begin{equation}\label{eq:latent_construction}
\begin{aligned}
    \mathbf{h}^m
    &= \operatorname{Encoder}(\tilde{h}^m),\\
    \mu^m
    &= \operatorname{Linear}_{\mu}(\mathbf{h}^m),
    \quad
    \sigma^m
    = \operatorname{Linear}_{\sigma}(\mathbf{h}^m),\\
    \tilde{z}^m
    &= \mu^m + \sigma^m \odot \epsilon,
    \quad
    \epsilon \sim \mathcal{N}(\mathbf{0},\mathbf{I}).
\end{aligned}
\end{equation}
where $\tilde{z}^m$ represents the sampled latent embedding of modality $m$ under simulated perturbation. Applying the same encoding and reparameterization procedure to the original unperturbed modality embedding $h^m$ yields the corresponding latent embedding $z^m$.

\noindent\textbf{Cross-Modal Alignment Loss.}
Following latent calibration via $\mathcal{L}_{SCL}$ (Eq.~\eqref{eq:SCL_loss}), the VAE decoder reconstructs the feature embeddings for each modality $m$. These reconstructed features are subsequently aggregated to synthesize the global joint embeddings $\hat{h}_S$ and $\hat{h}_T$ for the source and target MMKGs, respectively. To enforce cross-graph semantic consistency between these reconstructed representations, we minimize their alignment discrepancy via Mean Squared Error (MSE):
\begin{equation}\label{eq:cmal}
    \mathcal{L}_{CMAL}(\hat{h}_S, \hat{h}_T) = \text{MSE}(\hat{h}_S, \hat{h}_T).
\end{equation}

\subsection{Latent Semantic Calibration (LSC)}\label{sec:3.2}
To mitigate semantic inconsistencies arising from Context Imbalance (CI), we introduce a Latent Semantic Calibration (LSC) layer. While existing generative approaches~\cite{guo2023revisit} align features solely by forcing decoded reconstructions to approximate the original embeddings, they fail to regulate the latent space, overlooking structural discrepancies. LSC explicitly aligns the cross-graph latent distributions of equivalent entities, dynamically adjusting modality weights to mitigate suppression while constraining the reconstructed features to alleviate DM-induced overvaluation.

\noindent\textbf{Latent Embedding Extraction. }
To align and calibrate semantics across KGs, we extract calibrated latent embeddings from both masked and unmasked representations. Specifically, we apply a shared linear transformation followed by a hyperbolic tangent activation function to map the latent embeddings into a calibrated latent semantic space $\{\mathbf{z}^m, \tilde{\mathbf{z}}^m\} = \tanh(W_m \{z^m, \tilde{z}^m\} + b_m)$, where $\mathbf{z}^m$ and $\tilde{\mathbf{z}}^m$ denote the calibrated latent embeddings of the unmasked and masked modalities, respectively, with $W_m$ and $b_m$ as learnable parameters.

\noindent\textbf{Cross-Modal Embedding Fusion. }
To capture the varying semantic contributions of different modalities, we compute dynamic modality weights using a Multi-Layer Perceptron (MLP). The normalized weight for the calibrated latent embedding of a masked modality $m$ is defined as:
\begin{equation}\label{eq:modality_weight}
    \tilde{\mathbf{w}}^m = \frac{\text{MLP}(\tilde{\mathbf{z}}^m)}{\sum_{j \in \{v, g, r, a\}} \text{MLP}(\mathbf{z}^j)}
\end{equation}
Similarly, unmasked weights $\mathbf{w}^m$ are computed using $\mathbf{z}^m$. The final joint latent embedding $\tilde{\mathbf{z}}$ is constructed by fusing the calibrated embeddings $\tilde{\mathbf{z}} = \sum_{m \in \{v, g, r, a\}} \mathbf{z}^m \odot \mathbf{w}^m$, where $\mathbf{z}^m$ is replaced by the masked equivalent $\tilde{\mathbf{z}}^m$ when modality $m$ is missing.

\noindent\textbf{Cross-Modal Calibration Loss. }
To bridge the cross-KG semantic gaps induced by CI, we minimize the discrepancy between the masked joint latent representation of the source entity, $\tilde{\mathbf{z}}_S \in \mathcal{E}_S$, and the unmasked joint latent representation of the target entity, $\mathbf{z}_T \in \mathcal{E}_T$. Both embeddings are mapped through a ReLU activation and L1-normalized to construct valid probability distributions. We enforce distribution alignment across the latent space using a symmetric Kullback-Leibler (KL) divergence loss:
\begin{equation}\label{eq:SCL_loss}
    \mathcal{L}_{SCL}(\tilde{\mathbf{z}}_S, \mathbf{z}_T) = D_{KL}(\tilde{\mathbf{z}}_S \| \mathbf{z}_T) + D_{KL}(\mathbf{z}_T \| \tilde{\mathbf{z}}_S)
\end{equation}
where $\mathcal{L}_{SCL}$ aligns the high-level semantic distributions, thereby regulating the dynamic modality weights to mitigate neighborhood-driven suppression.

\noindent\textbf{Joint Learning Objective. }
In addition to $\mathcal{L}_{SCL}$ and $\mathcal{L}_{CMAL}$, ALMEA employs two standard VAE objectives for latent regularization and modality reconstruction, respectively:
\begin{equation}\label{eq:vae_losses}
\begin{aligned}
    \mathcal{L}_{DML} &= \frac{1}{2}\sum_{j=1}^{d} \left[ (\sigma_j^m)^2 + (\mu_j^m)^2 - \log\left((\sigma_j^m)^2\right) - 1 \right],\\
    \mathcal{L}_{URL} &= \operatorname{MSE} \left( \operatorname{Decoder}(\tilde{\mathbf{z}}^m), \mathbf{h}^m \right).
\end{aligned}
\end{equation}
The overall objective is:
\begin{equation}
\label{eq:joint_learning_objective}
\mathcal{L} = \lambda_1\mathcal{L}_{SCL} + \lambda_2\mathcal{L}_{DML} + \lambda_3\mathcal{L}_{URL} + \lambda_4\mathcal{L}_{CMAL}.
\end{equation}
Here, $\lambda_1,\lambda_2,\lambda_3,\lambda_4$ denotes the non-negative weights of the four objectives.

\subsection{Controlled Iterative Selection (CIS)}     
Under sparse supervision, the lack of sufficient equivalent seed pairs restricts the model's ability to robustly align latent semantic spaces. To overcome this, we propose Controlled Iterative Selection (CIS), which adaptively selects a highly representative and diverse subset of candidate pairs from the unlabeled pool under a constrained budget. CIS leverages the Dissimilarity-based Sparse Subset Selection (DS3) objective~\cite{elhamifar2015dissimilarity} as an active acquisition rule, progressively expanding the labeled seed set $L$ to iteratively refine LSC.

\noindent\textbf{Candidate Pool \( Q_{sorted} \). }
After obtaining semantically calibrated embeddings for all entities, the module iteratively refines embedding quality. 
Let $\mathcal{E}_S = \{\mathbf{h}_{S,1}, \dots, \mathbf{h}_{S,M}\}$ and $\mathcal{E}_T = \{\mathbf{h}_{T,1}, \dots, \mathbf{h}_{T,N}\}$ represent the joint embeddings of source and target entities. Since nearest-neighbor mappings across graphs are asymmetric, we construct a candidate set $Q$ of mutually high-confidence pairings using a symmetric mutual nearest-neighbor constraint:
\begin{equation}
\begin{aligned}
Q=\{(i,j)\mid\;& j=\arg\min_{j'} \|\mathbf{h}_{S,i} - \mathbf{h}_{T,j'}\|^2,\\
& i=\arg\min_{i'} \|\mathbf{h}_{T,j} - \mathbf{h}_{S,i'}\|^2\}
\end{aligned}
\label{eq:Q}
\end{equation}
where $j' \in \{1,\dots,N\}$ and $i' \in \{1,\dots,M\}$. We sort the candidate pairs in $Q$ in ascending order of their Euclidean distances to obtain the ranked pool $Q_{sorted} = \text{sort}(Q)$.

\noindent\textbf{The Representativeness Perspective. }
Let $Z \in \mathbb{R}^{K \times K}$ represent an indicator matrix where the element $Z[i,j] \in [0,1]$ denotes the probability that candidate pair $q_j \in Q_{sorted}$ is represented by candidate pair $q_i \in Q_{sorted}$, where . To select a small, optimal set of representative pairs, we promote row sparsity in $Z$ using the $L_{2,1}$-mixed norm $\|Z\|_{2,1} = \sum_{i=1}^{K} \|Z[i,:]\|_2$ where $\|Z[i,:]\|_2 > 0$ indicates that candidate $q_i$ is selected to represent other candidate pairs.

\noindent\textbf{Representative and Diversity-Aware Selection. }
We evaluate candidate pairs from two perspectives: pairwise representativeness within the candidate pool to prevent redundant selection and diversity to maximize the informational
coverage of the updated labeled set \(L\). These are defined as: 
\begin{equation}\label{eq:CIS_metrics}
\begin{aligned}
  A[i,j] &= \frac{1}{2} \left( \|\mathbf{h}_{S,i}-\mathbf{h}_{S,j}\|^2 + \|\mathbf{h}_{T,i}-\mathbf{h}_{T,j}\|^2 \right),\\
  C[i,l] &= \frac{1}{2} \left( \|\mathbf{h}_{S,i}-\mathbf{h}_{L^S_l}\|^2 + \|\mathbf{h}_{T,i}-\mathbf{h}_{L^T_l}\|^2 \right),\\
  D_{q_i} &= \sigma-(\sigma-1) \frac{\min_{l\in L}C[i,l]} {\max_{q_k\in Q}\min_{l\in L}C[k,l]}.
\end{aligned}
\end{equation}
Here, $(S_i, T_i) \in Q_{sorted}$ and $(L^S_j, L^T_j) \in L$. A smaller dissimilarity value $A[i,j]$ increases the probability of $q_j$ being represented by $q_i$. In contrast, a candidate farther from \(L\) receives a smaller diversity penalty \(D_{q_i}\) to increase its selection priority.

\noindent\textbf{Optimization. }
Incorporating the representativeness and diversity perspectives, the objective function is formulated as follows:
\begin{equation}\label{eq:original_optimise_problem}
    \min_{Z} \lambda \|DZ\|_{2,1} + \text{tr}(A^T Z) \quad \text{s.t.} \quad \mathbf{1}^T Z = \mathbf{1}^T, \quad Z \geq 0
\end{equation}
where $D = \text{diag}(D_{q_1}, \dots, D_{q_K})$ represents the diagonal diversity penalty matrix, $\lambda$ is a regularization parameter controlling sparsity, and $\text{tr}(\cdot)$ denotes the trace operator. 
We solve this convex optimization problem using ADMM. The alternating updates are detailed in Sections~\ref{sec:optimize formulation and derivation} and~\ref{sec:final} of the Supplementary Material. The overall workflow is summarized in Algorithm~\ref{alg:algorithm1}.

\begin{algorithm}[t]
    \caption{ALMEA: Controlled Iterative Selection for MMEA}
    \label{alg:algorithm1}
    \begin{algorithmic}[1]
        \STATE \textbf{Input:} Labeled set \(L\), unlabeled set \(U\), total epochs \(T_{\mathrm{total}}\), warm-up epochs \(T_{\mathrm{warm}}\), CIS interval \(T_{\mathrm{inter}}\), number of CIS rounds \(T_r\), maximum ADMM iterations \(T_{\mathrm{iter}}\), and confidence threshold \(\tau\)
        \STATE \textbf{Output:} Updated labeled set \(L\)

        \STATE Initialize model \(\mathbf{M}\) using \(L\)
        \STATE \(\mathcal{T}_{\mathrm{CIS}}
        \gets
        \{T_{\mathrm{warm}}+rT_{\mathrm{inter}}
        \mid r=0,\ldots,T_r-1\}\)

        \FOR{\(t=1\) to \(T_{\mathrm{total}}\)}
            \STATE Train \(\mathbf{M}\) on \(L\)

            \IF{\(t\in\mathcal{T}_{\mathrm{CIS}}\)}
                \STATE Construct \(Q_{\mathrm{sorted}}\) using Eq.~\ref{eq:Q}
                \STATE Construct \(A\), \(C\), and \(D\) using Eq.~\ref{eq:CIS_metrics}
                \STATE Solve Eq.~\ref{eq:original_optimise_problem} by ADMM within \(T_{\mathrm{iter}}\) iterations to obtain \(Z^*\)
                \STATE \(Q_{\mathrm{selected}}
                \gets
                \{q_i\in Q_{\mathrm{sorted}}
                \mid
                \|Z^*[i,:]\|_2>0,\,
                c(q_i)\geq\tau\}\)
                \STATE \(L\gets L\cup Q_{\mathrm{selected}}\)
                \STATE \(U\gets U\setminus Q_{\mathrm{selected}}\)
            \ENDIF
        \ENDFOR

        \STATE \textbf{return} \(L\)
    \end{algorithmic}
\end{algorithm}

%% file: 05-experiment.tex
\begin{table*}[ht]
  \centering
  \setlength\tabcolsep{0.95pt}
  \caption{Cross-MMKG alignment accuracy under 20\%, 50\%, and 80\% supervision. \textbf{Bold} represents two variants and the improvement, while \underline{underlining} indicates the strongest baseline; $*$ denotes results reproduced from released code, and the shade of each block represents the result of a paired $t$-test with ALMEA, based on five matched runs: \protect\colorbox{sigOneColor}{\strut significant}, \protect\colorbox{sigTwoColor}{\strut highly significant}, \protect\colorbox{sigThreeColor}{\strut extremely significant}, and \protect\colorbox{sigNSColor}{\strut not significant}. Improv.: A Comparison of ALMEA and SimDiff}
  
  \label{experiment_result}
  \resizebox{1\textwidth}{!}{
    \begin{tabular}{lccccccccc}
        \toprule
        & \multicolumn{3}{c}{FB15K-DB15K (20\%)} & \multicolumn{3}{c}{FB15K-DB15K (50\%)} & \multicolumn{3}{c}{FB15K-DB15K (80\%)} \\
        \cmidrule(r){2-4} \cmidrule(lr){5-7} \cmidrule(l){8-10}
        Methods & MRR(\%) & Hits@1(\%) & Hits@10(\%) & MRR(\%) & Hits@1(\%) & Hits@10(\%) & MRR(\%) & Hits@1(\%) & Hits@10(\%) \\
        \midrule
        TransE & 13.40 & \phantom{0}7.80 & 24.00 & 30.60 & 23.00 & 44.60 & 50.70 & 42.60 & 65.90 \\
        GCN-align & \phantom{0}8.70 & \phantom{0}5.30 & 17.40 & 29.30 & 22.60 & 43.50 & 47.20 & 41.40 & 63.50 \\
        SEA & 25.50 & 17.00 & 42.50 & 47.00 & 37.30 & 65.70 & 50.50 & 51.20 & 78.40 \\
        \midrule
        MMEA & 35.70 & 26.50 & 54.10 & 51.20 & 41.70 & 70.30 & 68.50 & 59.00 & 86.90 \\
        EVA & 18.48 & 11.90 & 31.60 & 26.14 & 18.06 & 42.32 & 35.34 & 25.98 & 53.48 \\
        \sigtwo{\strut MSNEA\(^*\)} & 37.89 {\scriptsize ± 0.4} & 28.74 {\scriptsize ± 0.3} & 55.78 {\scriptsize ± 0.3} & 62.85 {\scriptsize ± 6.2} & 54.76 {\scriptsize ± 7.5} & 75.56 {\scriptsize ± 6.2} & 78.97 {\scriptsize ± 5.9} & 73.44 {\scriptsize ± 6.7} & 88.53 {\scriptsize ± 4.1} \\
        \sigthree{\strut MCLEA\(^*\)} & 40.80 {\scriptsize ± 0.2} & 31.18 {\scriptsize ± 0.1} & 59.28 {\scriptsize ± 0.1} & 68.27 {\scriptsize ± 0.3} & 60.50 {\scriptsize ± 0.6} & 81.99 {\scriptsize ± 0.7} & 80.99 {\scriptsize ± 0.1} & 75.39 {\scriptsize ± 0.3} & 90.11 {\scriptsize ± 0.2} \\
        \sigtwo{\strut GEEA\(^*\)} & 40.23 {\scriptsize ± 0.1} & 30.12 {\scriptsize ± 0.2} & 60.34 {\scriptsize ± 0.2} & 64.84 {\scriptsize ± 0.2} & 55.99 {\scriptsize ± 0.1} & 81.01 {\scriptsize ± 0.3} & 79.46 {\scriptsize ± 0.7} & 73.51 {\scriptsize ± 0.1} & 89.45 {\scriptsize ± 0.7} \\
        \sigthree{\strut MEAformer\(^*\)} & 51.84 {\scriptsize ± 0.2} & 41.69 {\scriptsize ± 0.2} & 71.35 {\scriptsize ± 0.1} & 69.79 {\scriptsize ± 0.1} & 61.70 {\scriptsize ± 0.1} & 84.02 {\scriptsize ± 0.1} & 81.97 {\scriptsize ± 0.1} & 76.39 {\scriptsize ± 0.1} & 91.22 {\scriptsize ± 0.1} \\
        \sigone{\strut OTMEA\(^*\)} & 44.68 {\scriptsize ± 5.1} & 37.47 {\scriptsize ± 4.4} & 58.49 {\scriptsize ± 6.4} & 65.00 {\scriptsize ± 3.3} & 58.57 {\scriptsize ± 3.3} & 76.21 {\scriptsize ± 3.4} & 77.59 {\scriptsize ± 3.8} & 72.41 {\scriptsize ± 4.5} & 86.43 {\scriptsize ± 2.3} \\
        \sigthree{\strut SimDiff\(^*\)} & \underline{54.31} {\scriptsize ± 0.3} & \underline{44.39} {\scriptsize ± 0.3} & \underline{72.99} {\scriptsize ± 0.2} & \underline{71.56} {\scriptsize ± 0.1} & \underline{63.58} {\scriptsize ± 0.1} & \underline{85.42} {\scriptsize ± 0.1} & \underline{83.61} {\scriptsize ± 0.1} & \underline{78.74} {\scriptsize ± 0.1} & \underline{92.03} {\scriptsize ± 0.1} \\

        \midrule
        ALMEA$_{\text{w/o CIS}}$ & \textbf{56.00} {\scriptsize ± 0.1} & \textbf{46.81} {\scriptsize ± 0.1} & \textbf{73.52} {\scriptsize ± 0.2} & \textbf{74.20} {\scriptsize ± 0.2} & \textbf{67.41} {\scriptsize ± 0.2} & \textbf{86.31} {\scriptsize ± 0.4} & \textbf{84.83} {\scriptsize ± 0.0} & \textbf{80.45} {\scriptsize ± 0.2} & \textbf{93.28} {\scriptsize ± 0.1}\\

        ALMEA & \textbf{59.43} {\scriptsize ± 0.1} & \textbf{50.79} {\scriptsize ± 0.1} & \textbf{75.64} {\scriptsize ± 0.2} & \textbf{74.16} {\scriptsize ± 0.2} & \textbf{67.40} {\scriptsize ± 0.3} & \textbf{86.16} {\scriptsize ± 0.2} & \textbf{83.80} {\scriptsize ± 0.1} & \textbf{79.08} {\scriptsize ± 0.2} & \textbf{92.41} {\scriptsize ± 0.4}\\
        Improv. & \phantom{0}\textbf{5.12} {\scriptsize \phantom{± 0.0}} & \phantom{0}\textbf{6.40} {\scriptsize \phantom{± 0.0}} & \phantom{0}\textbf{2.65} {\scriptsize \phantom{± 0.0}} & \phantom{0}\textbf{2.60} {\scriptsize \phantom{± 0.0}} & \phantom{0}\textbf{3.82} {\scriptsize \phantom{± 0.0}} & \phantom{0}\textbf{0.74} {\scriptsize \phantom{± 0.0}} & \phantom{0}\textbf{0.19} {\scriptsize \phantom{± 0.0}} & \phantom{0}\textbf{0.34} {\scriptsize \phantom{± 0.0}} & \phantom{0}\textbf{0.38} {\scriptsize \phantom{± 0.0}} \\

    \end{tabular}}
    
\resizebox{1\textwidth}{!}{
    \begin{tabular}{lccccccccc}
        \toprule
        & \multicolumn{3}{c}{FB15K-YAGO15K (20\%)} & \multicolumn{3}{c}{FB15K-YAGO15K (50\%)} & \multicolumn{3}{c}{FB15K-YAGO15K (80\%)} \\
        \cmidrule(r){2-4} \cmidrule(lr){5-7} \cmidrule(l){8-10}
        Methods & MRR(\%) & Hits@1(\%) & Hits@10(\%) & MRR(\%) & Hits@1(\%) & Hits@10(\%) & MRR(\%) & Hits@1(\%) & Hits@10(\%) \\
        \midrule
        TransE & 11.20 & \phantom{0}6.40 & 20.30 & 26.20 & 19.70 & 38.20 & 46.30 & 39.20 & 59.50 \\
        GCN-align & 15.30 & \phantom{0}8.10 & 23.50 & 29.40 & 23.50 & 42.40 & 47.70 & 40.60 & 64.30 \\
        SEA & 21.80 & 14.10 & 37.10 & 38.80 & 29.40 & 57.70 & 60.50 & 51.40 & 77.30 \\
        \midrule
        MMEA & 31.70 & 23.40 & 48.00 & 48.60 & 40.30 & 64.50 & 68.20 & 59.80 & 83.90 \\
        EVA & 14.86 & 9.38 & 25.52 & 23.36 & 15.98 & 37.76 & 38.86 & 30.12 & 55.16 \\
        \sigtwo{\strut MSNEA\(^*\)} & 36.74 {\scriptsize ± 2.2} & 28.25 {\scriptsize ± 2.2} & 53.29 {\scriptsize ± 2.4} & 58.30 {\scriptsize ± 4.3} & 50.17 {\scriptsize ± 4.4} & 73.93 {\scriptsize ± 4.2} & 69.62 {\scriptsize ± 6.7} & 62.57 {\scriptsize ± 6.7} & 82.82 {\scriptsize ± 6.3} \\
        \sigthree{\strut MCLEA\(^*\)} & 36.48 {\scriptsize ± 0.1} & 28.35 {\scriptsize ± 0.1} & 51.76 {\scriptsize ± 0.2} & 59.36 {\scriptsize ± 0.1} & 50.64 {\scriptsize ± 0.2} & 74.40 {\scriptsize ± 0.7} & 76.59 {\scriptsize ± 0.2} & 70.61 {\scriptsize ± 0.1} & 86.90 {\scriptsize ± 0.3} \\
        \sigthree{\strut GEEA\(^*\)} & 33.65 {\scriptsize ± 0.1} & 25.38 {\scriptsize ± 0.1} & 49.77 {\scriptsize ± 0.0} & 57.59 {\scriptsize ± 0.1} & 49.43 {\scriptsize ± 0.0} & 72.65 {\scriptsize ± 0.3} & 74.31 {\scriptsize ± 0.1} & 67.59 {\scriptsize ± 0.1} & 86.01 {\scriptsize ± 0.1} \\
        \sigthree{\strut MEAformer\(^*\)} & 41.77 {\scriptsize ± 0.1} & 32.73 {\scriptsize ± 0.1} & 59.33 {\scriptsize ± 0.1} & 63.74 {\scriptsize ± 0.1} & 55.74 {\scriptsize ± 0.1} & 78.02 {\scriptsize ± 0.1} & 76.67 {\scriptsize ± 0.1} & 70.46 {\scriptsize ± 0.1} & 87.50 {\scriptsize ± 0.2} \\
        \sigone{\strut OTMEA\(^*\)} & 43.50 {\scriptsize ± 1.0} & 35.21 {\scriptsize ± 0.8} & 59.18 {\scriptsize ± 1.4} & 62.97 {\scriptsize ± 2.3} & 56.40 {\scriptsize ± 2.1} & 74.62 {\scriptsize ± 2.8} & 71.24 {\scriptsize ± 8.7} & 65.12 {\scriptsize ± 7.9} & 79.78 {\scriptsize ± 8.7} \\
        \sigthree{\strut SimDiff\(^*\)} & \underline{45.72} {\scriptsize ± 0.1} & \underline{36.61} {\scriptsize ± 0.1} & \underline{62.96} {\scriptsize ± 0.1} & \underline{64.35} {\scriptsize ± 0.1} & \underline{56.25} {\scriptsize ± 0.1} & \underline{78.69} {\scriptsize ± 0.1} & \underline{77.08} {\scriptsize ± 0.1} & \underline{71.30} {\scriptsize ± 0.1} & \underline{87.06} {\scriptsize ± 0.1} \\
        \midrule
        ALMEA$_{\text{w/o CIS}}$ & \textbf{46.09} {\scriptsize ± 0.2} & \textbf{36.89} {\scriptsize ± 0.2} & \textbf{63.81} {\scriptsize ± 0.2} & \textbf{67.21} {\scriptsize ± 0.1} & \textbf{59.61} {\scriptsize ± 0.1} & \textbf{81.40} {\scriptsize ± 0.3} & \textbf{80.10} {\scriptsize ± 0.2} & \textbf{74.48} {\scriptsize ± 0.1} & \textbf{90.16} {\scriptsize ± 0.2} \\

        ALMEA & \textbf{50.92} {\scriptsize ± 0.1} & \textbf{42.11} {\scriptsize ± 0.1} & \textbf{67.70} {\scriptsize ± 0.1} & \textbf{68.77} {\scriptsize ± 0.1} & \textbf{61.33} {\scriptsize ± 0.1} & \textbf{82.50} {\scriptsize ± 0.2} & \textbf{78.34} {\scriptsize ± 0.4} & \textbf{72.27} {\scriptsize ± 0.5} & \textbf{88.95} {\scriptsize ± 0.5} \\
        Improve. & \phantom{0}\textbf{5.20} {\scriptsize \phantom{± 0.0}} & \phantom{0}\textbf{5.50} {\scriptsize \phantom{± 0.0}} & \phantom{0}\textbf{4.74} {\scriptsize \phantom{± 0.0}} & \phantom{0}\textbf{4.42} {\scriptsize \phantom{± 0.0}} & \phantom{0}\textbf{5.08} {\scriptsize \phantom{± 0.0}} & \phantom{0}\textbf{3.81} {\scriptsize \phantom{± 0.0}} & \phantom{0}\textbf{1.26} {\scriptsize \phantom{± 0.0}} & \phantom{0}\textbf{0.97} {\scriptsize \phantom{± 0.0}} & \phantom{0}\textbf{1.89} {\scriptsize \phantom{± 0.0}} \\

        \bottomrule
    \end{tabular}}
\end{table*}

\begin{table*}[ht]
  \centering
  \setlength\tabcolsep{0.95pt}
  \caption{Multilingual MMKGs Alignment Accuracy Comparison. The results comparison on three bilingual datasets in different settings: without using Surface Forms (w/o SF), with Surface Forms (w/ SF). Under the unsupervised (unsup) setting, ALMEA-V and ALMEA-N use visual and name similarities, respectively, to initialize the alignment space. Bold values highlight the ALMEA variants rather than the column-wise best results.}
  \label{experiment_result_2}
  \resizebox{1\textwidth}{!}{
    \begin{tabular}{llccccccccc}
        \toprule
        Setting & Methods 
        & \multicolumn{3}{c}{$\text{DBP15K}_{ZH-EN}$} 
        & \multicolumn{3}{c}{$\text{DBP15K}_{JA-EN}$} 
        & \multicolumn{3}{c}{$\text{DBP15K}_{FR-EN}$} \\
        \cmidrule(r){3-5} \cmidrule(lr){6-8} \cmidrule(l){9-11}
        & 
        & MRR(\%) & Hits@1(\%) & Hits@10(\%) 
        & MRR(\%) & Hits@1(\%) & Hits@10(\%) 
        & MRR(\%) & Hits@1(\%) & Hits@10(\%) \\
        \midrule
        \multirow{6}{*}{w/o SF} 
        & BootEA & 70.3 & 62.9 & 84.7 & 70.1 & 62.2 & 85.4 & 73.1 & 65.3 & 87.4 \\
        & EVA    & 81.4 & 76.1 & 90.7 & 81.7 & 76.2 & 91.3 & 84.7 & 79.3 & 94.2 \\
        & MCLEA  & 86.5 & 81.6 & 94.8 & 86.5 & 81.2 & 95.2 & 88.5 & 83.4 & 97.5 \\
        & SimDiff  & 87.7 & 82.9 & 96.3 & 88.3 & 83.3 & 96.6 & 90.5 & 86.1 & 98.0 \\
        & ALMEA$_{\text{w/o CIS}}$ & \textbf{81.8} & \textbf{74.7} & \textbf{94.4} & \textbf{81.3} & \textbf{74.5} & \textbf{93.3} & \textbf{82.7} & \textbf{75.9} & \textbf{94.7} \\
        & ALMEA  & \textbf{88.3} & \textbf{83.7} & \textbf{96.6} & \textbf{89.0} & \textbf{84.5} & \textbf{97.4} & \textbf{91.0} & \textbf{86.9} & \textbf{98.6} \\
        \midrule
        \multirow{5}{*}{w/ SF} 
        & AttrGNN & 82.9 & 77.7 & 92.0 & 81.6 & 76.3 & 90.9 & 95.9 & 94.2 & 98.7 \\
        & MCLEA    & 98.1 & 97.2 & 99.6 & 99.1 & 98.6 & 99.9 & 99.8 & 99.7 & 100.0 \\
        & SimDiff  & 98.3 & 97.4 & 99.8 & 99.3 & 98.9 & 99.9 & 99.8 & 99.7 & 100.0\\
        & ALMEA$_{\text{w/o CIS}}$ & \textbf{94.7} & \textbf{92.3} & \textbf{98.6} & \textbf{97.5} & \textbf{96.2} & \textbf{99.5} & \textbf{93.3} & \textbf{88.9} & \textbf{99.9} \\
        & ALMEA  & \textbf{98.4} & \textbf{97.4} & \textbf{99.9} & \textbf{99.4} & \textbf{99.0} & \textbf{99.9} & \textbf{99.8} & \textbf{99.9} & \textbf{100.0} \\
        \midrule
        \multirow{8}{*}{unsup} 
        & MCLEA-V & 97.4 & 95.9 & 99.5 & 98.7 & 97.7 & 99.9 & 99.4 & 99.0 & 100.0 \\
        & MCLEA-N    & 97.1 & 95.5 & 99.4 & 98.9 & 97.9 & 99.9 & 99.7 & 99.5 & 100.0 \\
        & SimDiff-N  & 97.3 & 95.6 & 99.5 & 98.9 & 98.3 & 99.9 & 99.3 & 98.7 & 100.0\\
        & SimDiff-V  & 97.6 & 96.4 & 99.6 & 99.0 & 98.4 & 99.9 & 99.7 & 99.5 & 100.0\\
        & ALMEA-N$_{\text{w/o CIS}}$ & \textbf{81.8} & \textbf{77.5} & \textbf{89.1} & \textbf{86.5} & \textbf{82.8} & \textbf{92.6} & \textbf{92.2} & \textbf{89.5} & \textbf{96.9} \\
        & ALMEA-N  & \textbf{97.3} & \textbf{95.4} & \textbf{99.5} & \textbf{98.9} & \textbf{96.3} & \textbf{99.9} & \textbf{99.8} & \textbf{99.6} & \textbf{100.0} \\
        & ALMEA-V$_{\text{w/o CIS}}$ & \textbf{82.5} & \textbf{78.5} & \textbf{89.4} & \textbf{89.1} & \textbf{85.9} & \textbf{94.5} & \textbf{94.6} & \textbf{92.6} & \textbf{97.9} \\
        & ALMEA-V  & \textbf{97.4} & \textbf{95.9} & \textbf{99.5} & \textbf{99.2} & \textbf{98.4} & \textbf{99.9} & \textbf{99.8} & \textbf{99.7} & \textbf{100.0} \\
        \bottomrule
    \end{tabular}}
\end{table*}

\section{Experiments}\label{sec:exp}
In this section, we evaluate the performance of the proposed ALMEA framework. We first present the experimental setup, followed by a comparative analysis against state-of-the-art baselines under various resource constraints, an analysis of time complexity, a comprehensive ablation study, and an in-depth investigation of the Controlled Iterative Selection (CIS) mechanism under limited supervision.

\subsection{Experimental Settings} \label{sec:setting}

\par\noindent \textbf{Datasets.} \label{sec:dataset}
We evaluate the MMEA task on two standard cross-KG benchmarks: FB15K-DB15K and FB15K-YAGO15K~\cite{Garcia-DuranN18}, which contain relational, visual, textual, and numerical attributes. To evaluate the generalizability of our model beyond English-only cross-KG settings, we also conduct experiments on the multilingual DBP15K benchmarks ($\text{DBP15K}_{ZH-EN}$, $\text{DBP15K}_{JA-EN}$, $\text{DBP15K}_{FR-EN}$)~\cite{sun2017cross}. The ground-truth (GT) equivalent entity pairs are split into 20\%, 50\%, and 80\% training ratios to represent low-to-high resource scenarios.

\par\noindent \textbf{Baselines.} \label{sec:baseline}
We compare ALMEA against two categories of established models: (i) unimodal entity alignment (EA) methods, including TransE~\cite{bordes2013translating}, GCN-Align~\cite{wang2018cross}, and SEA~\cite{pei2019semi}; and (ii) multimodal entity alignment (MMEA) methods, including MMEA~\cite{chen2020mmea}, EVA~\cite{liu2021visual}, MSNEA~\cite{chen2022multi}, MCLEA~\cite{lin2022multi}, GEEA~\cite{guo2023revisit}, MEAformer~\cite{chen2023meaformer}, OTMEA~\cite{wang2025otmea}, and SimDiff~\cite{li2024simdiff}. 

For the multilingual DBP15K benchmarks, we evaluate under three paradigms: without Surface Forms (w/o SF), with Surface Forms (w/SF), and unsupervised (unsup). Under the unsupervised paradigm, we implement two variants: \textbf{ALMEA-V}, which utilizes visual similarity to initialize the alignment space following EVA~\cite{liu2021visual}, and \textbf{ALMEA-N}, which leverages name similarity as the initial supervisory signal following MCLEA~\cite{lin2022multi}.

\par\noindent \textbf{Evaluation Metrics.} \label{sec:metric}
Alignment performance is quantitatively measured using standard ranking metrics, specifically Hits@$n$ (where $n \in \{1, 10\}$) and Mean Reciprocal Rank (MRR). Higher values in Hits@$n$ and MRR represent superior alignment accuracy and rank retrieval.

\par\noindent \textbf{Implementation Details.} \label{sec:impl}
We implement a GAT-based network to encode graph structural features, GloVe-6B to embed textual attributes, and a VGG16 architecture to process visual features, setting the projection space dimension to $d=300$ for all modalities. Optimization is performed using the Adam optimizer with a batch size of 3,500. Detailed hyperparameter configurations and hardware details are provided in Supplementary Material~\ref{sec:more exp details}.

\subsection{Overall Performance} \label{sec:rq1}
To reduce experimental bias, we report the mean and standard deviation computed over five independent trials for MSNEA, MCLEA, GEEA, MEAformer, OTMEA, and SimDiff. To further evaluate the efficacy of \textbf{ALMEA}, we conducted paired t-tests~\cite{student1908probable,demvsar2006statistical} against those models we mentioned before. As shown in Table~\ref{experiment_result}, the performance improvements achieved by ALMEA are statistically significant in most cases, as indicated by the color coding. The overall performance comparison on FB15K-DB15K (top) and FB15K-YAGO15K (bottom). Here, $\text{ALMEA}_{\text{w/o CIS}}$ denotes our model optimized without iterative learning, and $\text{ALMEA}$ denotes the complete model incorporating CIS.

The experimental results show that ALMEA consistently outperforms all baseline models across all training scales (20\%, 50\%, 80\%) and evaluation metrics on both cross-KG datasets. Even without CIS, $\text{ALMEA}_{\text{w/o CIS}}$ improves the MRR metric by 1.22--2.64\% on FB15K-DB15K and 0.37--3.02\% on FB15K-YAGO15K compared with the strongest baselines. Under 20\% supervision, incorporating CIS further enlarges these gains, whereas its incremental benefit diminishes as the available supervision increases. This indicates that coupling CMI (low-level feature imputation) with LSC (latent semantic calibration) within a bidirectional feedback loop successfully mitigates the semantic degradation caused by the overvaluation of missing features and context-driven modality suppression.

Furthermore, we observe that ALMEA achieves larger performance gains on FB15K-YAGO15K (e.g., up to 5.50\% in Hits@1 at 20\% training) than on FB15K-DB15K. This pattern is consistent with the higher visual-modality missing rate of FB15K-YAGO15K (Figure~\ref{fig:mainfig}, middle), suggesting that our VAE-based imputation remains robust under severe Detail Missing (DM) settings. As the supervision ratio increases, we observe diminishing returns in performance gains (gains of 3.91\%, 0.70\%, and $-1.41\%$ for training ratios of 20\%, 50\%, and 80\%, respectively), highlighting the particular utility of the CIS module in sparse-supervision (low-resource) scenarios.

For the multilingual DBP15K benchmarks, Table~\ref{experiment_result_2} shows that ALMEA demonstrably outperforms the state-of-the-art across all settings, achieving an average increase of 9.97\% in Hits@1 under the low-resource setting (30\% training ratio). Since DBP15K features richer, higher-quality attribute modalities than the cross-KG benchmarks, the CIS module can leverage these dense features to construct a highly representative and diverse labeled set, thereby further confirming its efficacy in multilingual contexts.

\subsection{Time Complexity Analysis}
\begin{table}[t]
\caption{Training efficiency. A$^{-}$/A$^{+}$ denote ALMEA without/with CIS; higher Iter/s indicates faster training.}
\label{tab:efficiency}
\centering
\setlength{\tabcolsep}{2pt}
\renewcommand{\arraystretch}{1.05}
\resizebox{\columnwidth}{!}{%
\begin{tabular}{@{}lccccccc@{}}
\toprule
& MCLEA & GEEA & MEAformer & OTMEA & SimDiff
& \textbf{A$^{-}$} & \textbf{A$^{+}$} \\
\midrule
Params (M)
& 9.06 & 47.46 & 10.58 & 11.02 & 10.31
& \textbf{14.34} & \textbf{14.34} \\
Iter/s $\uparrow$
& 3.94 & 0.73 & 0.74 & 0.69 & 0.36
& \textbf{1.28} & \textbf{1.12} \\
\bottomrule
\end{tabular}%
}
\end{table}
As shown in Table~\ref{tab:efficiency}, although ALMEA introduces additional parameters compared to some baselines (e.g., approximately 4.0M more parameters than the diffusion-based SimDiff), it exhibits superior training efficiency, running at 1.12 iterations per second (iter/s). 

The computational complexity of our CIS module is formulated as $O(M^2 d + (M^U)^2 + K^2 d + K M^L d + K M^L + T_{\text{iter}} K^2)$, where $M$ is the total number of entities, $M^L$ and $M^U$ are the sizes of the labeled and unlabeled subsets respectively, $K$ represents the candidate pool size ($K \ll M$), and $T_{\text{iter}}$ is the number of ADMM optimization iterations. Because $K$ and $T_{\text{iter}}$ are small, user-controlled hyperparameters in practice, the asymptotic complexity is upper-bounded by $O(M^2 d + M^2 + M d + M) = O(M^2 d)$. Consequently, our CIS framework scales quadratically with the entity scale $M$ rather than exponentially, making it computationally feasible for large KGs. A detailed complexity derivation is provided in Supplementary Material~\ref{sec:derivation of time complexity}.

\subsection{Ablation Study} \label{sec:rq3}
To dissect the semantic and structural impact of each algorithmic design in ALMEA, we conduct ablation studies. The VAE-based CMI and LSC modules are evaluated under an 80\% supervision setting to isolate representation learning effects, while the CIS module is evaluated under a 20\% supervision setting to assess low-resource efficacy. The results are summarized in Table~\ref{tab:ablation_study}.

\par\noindent \textbf{Impact of DM Disturbance Simulation (DMDS).}
Removing the proactive masking perturbation step (w/o DMDS) degrades the average MRR by 1.85\% across both datasets. This decrease confirms that intentionally training the VAE on severe, simulated missingness helps the model build robust, noise-resilient modality embeddings, validating its effectiveness in mitigating overvaluation.

\par\noindent \textbf{Impact of Latent Semantic Calibration (LSC).}
Disabling LSC (w/o LSC) degrades MRR by 3.10\% on FB15K-DB15K and 3.04\% on FB15K-YAGO15K. This performance drop highlights the key role LSC plays in aligning disparate latent semantics and preventing contextual imbalances from suppressing weaker modal features. This conclusion is further supported by ablation experiments on the loss layers: Removing $\mathcal{L}_{SCL}$ had the greatest impact on both datasets, while removing $\mathcal{L}_{CMAL}$ also degraded performance, demonstrating that it affects cross-knowledge-graph alignment; in contrast, removing the standard VAE loss ($\mathcal{L}_{DML}$ or $\mathcal{L}_{URL}$) primarily affects the regularization of modality reconstruction.
\par\noindent \textbf{Impact of Modality Removal.}
Removing individual modalities reveals that the relational modality (w/o Relation) is the most critical feature, and its omission leads to a sharp performance drop (e.g., degrading MRR by up to 4.14\% on FB15K-YAGO15K). This is because relational contexts, particularly multi-to-many properties, provide key topological constraints. Conversely, removing the visual modality (w/o Visual) results in the smallest degradation, which is attributed to the fact that single-image features provide weaker structural signals than relation or attribute neighborhoods.

\par
\begin{figure}[t]
\centering
\subcaptionbox{w/o diversity score $D$}{\includegraphics[width=0.48\columnwidth]{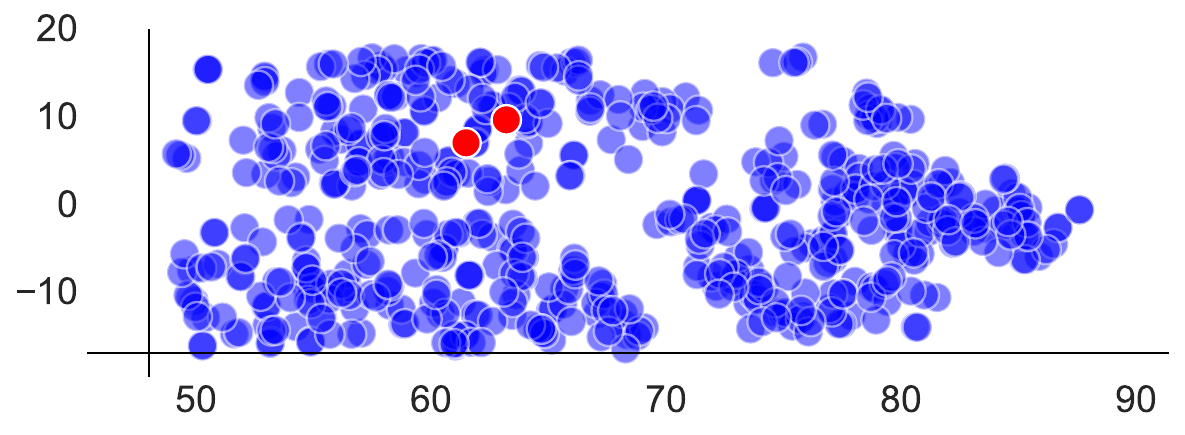}}
\hfill
\subcaptionbox{w diversity score $D$}{\includegraphics[width=0.48\columnwidth]{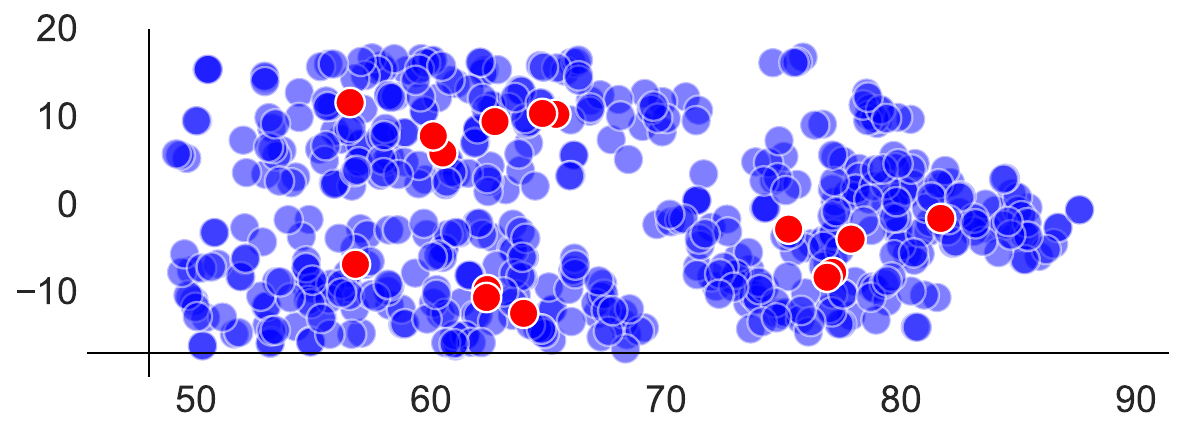}}

\caption{Effectiveness of diversity in iterative learning. t-SNE plots show the sub-dataset (blue) and the CIS-selected subset (red), comparing selections without and with diversity score $D$.} 
\label{fig:AL_analysis_for_RQ2}
\end{figure}

\noindent \textbf{Impact of Controlled Iterative Selection (CIS).}
Under low-resource conditions, removing the diversity score (w/o $D$) severely degrades performance, dropping Hits@1 by 6.09\% on FB15K-DB15K and 6.29\% on FB15K-YAGO15K. As illustrated in the t-SNE projections in Figure~\ref{fig:AL_analysis_for_RQ2}(a) and (b), omitting $D$ forces the iterative selection process to select candidate pairs concentrated in narrow high-density regions of the embedding space, resulting in spatial redundancy. In contrast, incorporating the diversity penalty $D$ guides the sparsification vector to cover structurally novel, diverse neighborhoods, which maximizes information gain under a restricted annotation budget.

\begin{table}[t]
    \centering
    \setlength\tabcolsep{1.2pt}
    \renewcommand{\arraystretch}{1.05}
    \caption{Ablation study. All metrics are reported in \%. Results are grouped into ALMEA$_{\text{w/o CIS}}$ and ALMEA$_{\text{w/ CIS}}$. ALMEA$_{\text{w/o CIS}}$ is tested under 80\% supervision, where ``Mod.'', ``Loss'', and ``Mdl.'' denote module-, loss-, and modality-level ablations, respectively. ALMEA$_{\text{w/ CIS}}$ is tested under 20\% supervision. DMDS denotes DM Disturbance Simulation.}
    \label{tab:ablation_study}
    \small
    \begin{tabular}{llcccccc}
        \toprule
         & Variants 
        & \multicolumn{3}{c}{FB15K-DB15K} 
        & \multicolumn{3}{c}{FB15K-YAGO15K} \\
        \cmidrule(lr){3-5} \cmidrule(lr){6-8}
        & 
        & MRR & Hits@1 & Hits@10  
        & MRR & Hits@1 & Hits@10 \\
        \midrule

        & ALMEA$_{\text{w/o CIS}}$ 
        & \textbf{84.83} & \textbf{80.45} & \textbf{93.28} 
        & \textbf{80.10} & \textbf{74.48} & \textbf{90.16} \\
        \midrule
        \multirow{2}{*}{\rotatebox{90}{Mod.}}
        & w/o DMDS
        & 82.80 & 77.13 & 90.94 
        & 78.43 & 72.45 & 88.83 \\
        & w/o LSC 
        & 81.73 & 76.41 & 89.89 
        & 77.06 & 71.21 & 87.48 \\
        \midrule
        \multirow{4}{*}{\rotatebox{90}{Loss}}
        & w/o $\mathcal{L}_{DML}$ 
        & 82.96 & 77.80 & 91.36 
        & 78.33 & 72.30 & 88.86 \\
        & w/o $\mathcal{L}_{URL}$ 
        & 83.46 & 78.50 & 91.51 
        & 78.56 & 72.67 & 88.99 \\
        & w/o $\mathcal{L}_{SCL}$ 
        & 82.13 & 76.83 & 90.24 
        & 77.40 & 71.53 & 87.53 \\
        & w/o $\mathcal{L}_{CMAL}$ 
        & 83.43 & 78.25 & 91.38 
        & 78.33 & 72.45 & 88.72 \\
        \midrule
        \multirow{3}{*}{\rotatebox{90}{Mdl.}}
        & w/o Relation 
        & 80.93 & 75.24 & 90.34 
        & 75.96 & 68.61 & 87.84 \\
        & w/o Attribute 
        & 81.90 & 76.24 & 91.17 
        & 76.66 & 70.03 & 88.00 \\
        & w/o Visual 
        & 83.56 & 78.58 & 91.78 
        & 78.46 & 72.27 & 89.01 \\
        
        \midrule

        & ALMEA$_{\text{w/ CIS}}$ 
        & \textbf{59.43} & \textbf{50.79} & \textbf{75.64} 
        & \textbf{50.92} & \textbf{42.11} & \textbf{67.70} \\
        \midrule
        \multirow{2}{*}{\rotatebox{90}{Mod.}}
        & w/o $Q_{sorted}$ 
        & 57.00 & 48.07 & 73.85 
        & 47.16 & 38.70 & 63.67 \\
        & w/o $D$ 
        & 54.30 & 44.70 & 72.55 
        & 44.86 & 35.82 & 62.35 \\
        
        \bottomrule
    \end{tabular}
\end{table}
\subsection{Quality of Imputed Modalities}
For each imputed modality, we retrieve its nearest neighbors retained under the same similarity criterion in the corresponding embedding space and compare them with the ground-truth modal information. Reconstructing missing modalities reflects the restoration of low-level details, while capturing high-level semantic distributions reflects the ability to generalize. Therefore, we compare ALMEA with GEEA, which also uses a VAE for modal reconstruction. Table~\ref{RQ4:perth airport} shows representative examples. ALMEA can recover more low-level details from neighborhood and attribute information. Furthermore, it preserves more neighborhood entities and attributes related to potential high-level semantics. These results indicate that ALMEA can reconstruct richer modality details, thereby reducing the loss of local information associated with overvaluation. For more imputed cases, see~\ref{Quality of Synthesized Modality}.
\begin{table}[t]
\caption{Qualitative comparison of imputed modalities for Perth Airport on FB15K. For each imputed modality, the nearest neighbors are retrieved in the embedding space. \textcolor{blue}{Blue} text denotes retrieved items that exactly match the ground truth, while \underline{underlined} text denotes additional non-ground-truth neighbors returned by KNN.}
\label{RQ4:perth airport}
\centering
\small
\setlength{\tabcolsep}{4pt}
\renewcommand{\arraystretch}{1.08}
\begin{tabular}{@{}>{\raggedright\arraybackslash}p{0.18\columnwidth}>{\raggedright\arraybackslash}p{0.76\columnwidth}@{}}
\toprule
\textbf{Source} & \textbf{Modal information} \\
\midrule
\multicolumn{2}{c}{\textbf{Neighborhood}} \\
\midrule
Ground Truth & Qantas, Jakarta, Cairns, Darwin\_Northern\_Territory, Canberra, Brisbane \\

ALMEA & \textcolor{blue}{Qantas}, \textcolor{blue}{Jakarta}, \textcolor{blue}{Cairns}, \textcolor{blue}{Darwin\_Northern\_Territory}, \textcolor{blue}{Canberra}, \textcolor{blue}{Brisbane}, \underline{Nine\_Network}, \underline{Network\_Ten}, \underline{Keith\_Urban}, \underline{Australia\_(2008\_film)}, \underline{Adelaide}, \underline{University\_of\_Adelaide} \\

GEEA & \textcolor{blue}{Brisbane}, \textcolor{blue}{Qantas}, \textcolor{blue}{Cairns}, \underline{Nine\_Network}, \underline{Network\_Ten}, \underline{University\_of\_Adelaide} \\

\midrule
\multicolumn{2}{c}{\textbf{Attribute}} \\
\midrule
Ground Truth & number\_of\_runways, longitude, mean\_elevation, latitude \\

ALMEA & \textcolor{blue}{number\_of\_runways}, \textcolor{blue}{longitude}, \textcolor{blue}{mean\_elevation}, \textcolor{blue}{latitude}, \underline{wasCreatedOnDate}, \underline{formationYear} \\

GEEA & \textcolor{blue}{number\_of\_runways}, \textcolor{blue}{longitude}, \textcolor{blue}{mean\_elevation}, \textcolor{blue}{latitude}, \underline{formationYear} \\
\bottomrule
\end{tabular}
\end{table}

\subsection{CIS under Limited Supervision}
\begin{figure*}[ht]
\centering

\begin{subfigure}[t]{0.31\textwidth}
    \centering
    \includegraphics[width=\textwidth]{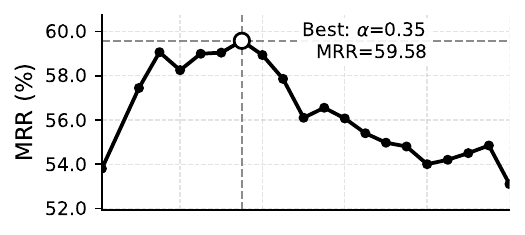}
    \caption{FB15K-DB15K (20\%)}
\end{subfigure}
\begin{subfigure}[t]{0.31\textwidth}
    \centering
    \includegraphics[width=\textwidth]{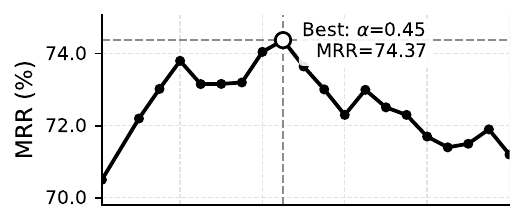}
    \caption{FB15K-DB15K (50\%)}
\end{subfigure}
\begin{subfigure}[t]{0.31\textwidth}
    \centering
    \includegraphics[width=\textwidth]{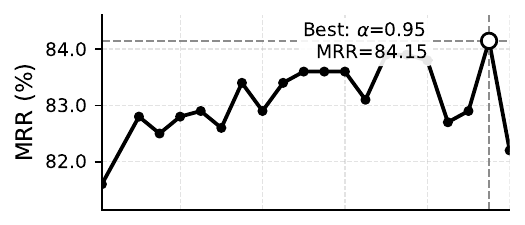}
    \caption{FB15K-DB15K (80\%)}
\end{subfigure}

\begin{subfigure}[t]{0.31\textwidth}
    \centering
    \includegraphics[width=\textwidth]{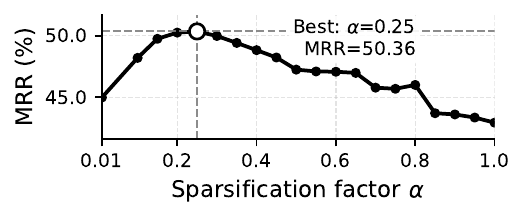}
    \caption{FB15K-YAGO15K (20\%)}
\end{subfigure}
\begin{subfigure}[t]{0.31\textwidth}
    \centering
    \includegraphics[width=\textwidth]{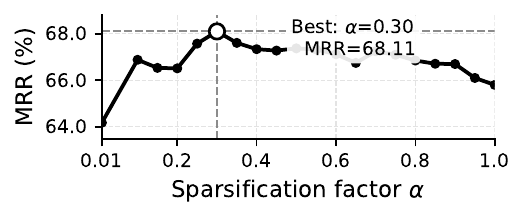}
    \caption{FB15K-YAGO15K (50\%)}
\end{subfigure}
\begin{subfigure}[t]{0.31\textwidth}
    \centering
    \includegraphics[width=\textwidth]{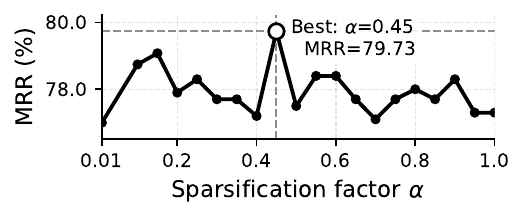}
    \caption{FB15K-YAGO15K (80\%)}
\end{subfigure}

\caption{Sensitivity analysis of the sparsification factor $\alpha$.}
\label{fig:alpha_two_datasets}
\end{figure*}
\begin{figure*}[ht]
\centering 

\begin{subfigure}[t]{0.31\textwidth}
    \centering
    \includegraphics[width=\textwidth]{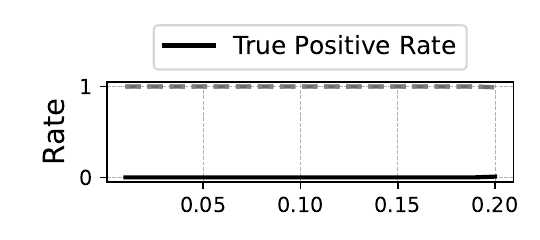}
    \caption{w/o $D$, \( \alpha = 0.1\)}
\end{subfigure}
\begin{subfigure}[t]{0.31\textwidth}
    \centering
    \includegraphics[width=\textwidth]{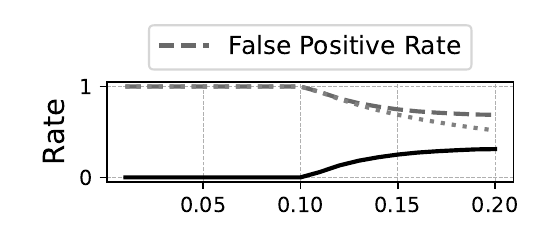}
    \caption{w/o $D$, \( \alpha = 0.35\)}
\end{subfigure}
\begin{subfigure}[t]{0.31\textwidth}
    \centering
    \includegraphics[width=\textwidth]{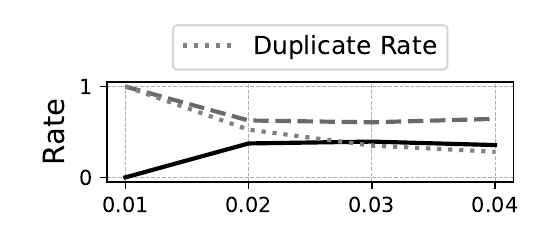}
    \caption{w/o $D$, \( \alpha = 1\)}
\end{subfigure}

\begin{subfigure}[t]{0.31\textwidth}
    \centering
    \includegraphics[width=\textwidth]{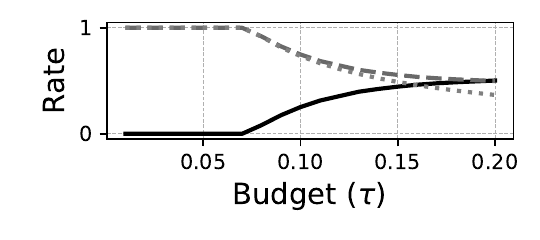}
    \caption{w/ $D$, \( \alpha = 0.1\)}
\end{subfigure}
\begin{subfigure}[t]{0.31\textwidth}
    \centering
    \includegraphics[width=\textwidth]{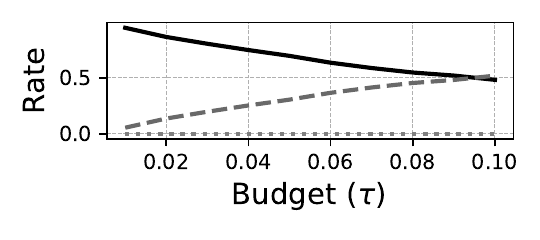}
    \caption{w/ $D$, \( \alpha = 0.35\)}
\end{subfigure}
\begin{subfigure}[t]{0.31\textwidth}
    \centering
    \includegraphics[width=\textwidth]{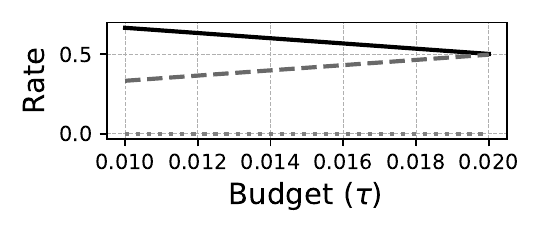}
    \caption{w/ $D$, \( \alpha = 1\)}
\end{subfigure}

    \caption{Candidate Pairs Quality Analysis of CIS. True positive, false positive, and duplication rates are shown over $\tau \in [0,0.2]$ for $\alpha \in \{0.1, 0.35, 1\}$, without $D$ ((a)--(c)) and with $D$ ((d)--(f)).}
\label{fig:AL analysis TPR}
\end{figure*}

\begin{figure*}[ht]
    \centering
    \includegraphics[width=1\linewidth]{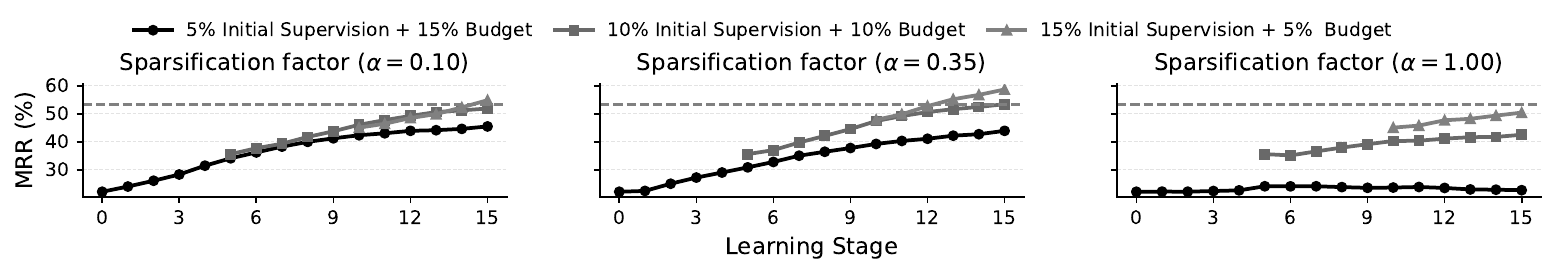}
    \caption{Downstream Gains of CIS in Low-Resource Settings.}\label{fig:result of RQ2}
\end{figure*}

To evaluate how CIS mitigates context imbalance and data missingness under restricted annotation budgets, we analyze its behavior across three aspects: parameter sensitivity, candidate selection quality, and downstream alignment accuracy.

\par\noindent \textbf{Sparsification Sensitivity of CIS.} \label{sec:sensitivity alpha}
As shown in Figure~\ref{fig:alpha_two_datasets}, we analyze alignment sensitivity across various sparsification factors $\alpha \in [0.01, 1]$. We observe two key trends. First, CIS is most effective under sparse low-resource settings, specifically at 20\% supervision (Figures~\ref{fig:alpha_two_datasets}(a), (d)), where selecting highly representative pairs significantly accelerates convergence. At 80\% supervision (Figures~\ref{fig:alpha_two_datasets}(c), (f)), the performance curves flatten because the labeled pool already contains sufficient structural information, validating the design of CIS for resource-constrained contexts. 

Second, the optimal sparsification factor increases with supervision: for FB15K-DB15K, the best-performing factor shifts from $\alpha=0.35$ to $\alpha=0.95$ as the training ratio increases from 20\% to 80\%. This indicates that as more training data becomes available, the optimization benefit shifts toward stronger sparsification (larger $\alpha$) to filter out redundant candidate pairs.

\par\noindent \textbf{Candidate Pairs Quality Analysis of CIS.} \label{sec:pair quality analysis}
We evaluate the quality of the selected candidate pairs by monitoring the True Positive Rate (TPR), False Positive Rate (FPR), and duplicate rate (the ratio of selected pairs that are already present in the existing labeled set). Figure~\ref{fig:AL analysis TPR} illustrates these metrics across selection thresholds $\tau \in [0, 0.2]$ for $\alpha \in \{0.1, 0.35, 1\}$, with and without the diversity score $D$.

As shown in Figures~\ref{fig:AL analysis TPR}(a)--(c), without incorporating the diversity score $D$, the selection optimization is driven entirely by representativeness ($Z$). When $\alpha = 0.35$, although the model successfully identifies correct candidate pairs (high TPR), the duplicate rate rises significantly because the selection algorithm has no directional constraint to avoid existing seed pairs. At $\alpha = 1$, the excessive sparsification constraint limits the selection pool, utilizing only 4\% of the total available budget.

Conversely, as shown in Figures~\ref{fig:AL analysis TPR}(d)--(f), incorporating the diversity score $D$ enforces a structural novelty constraint, effectively keeping the duplicate rate at 0\% across all configurations. For $\alpha = 0.35$ and $\alpha = 1$, the diversity matrix successfully restricts selection to novel areas of the embedding space, utilizing 10\% and 2\% of the candidate budget, respectively. Restricting the selected subset to the top 1\% of candidate pairs per iteration yields the highest performance, confirming the importance of balancing representativeness with spatial diversity.

\begin{table}[t]
\caption{CIS selection strategies at 20\%, 50\%, and 80\% supervision. DB and YG denote FB15K-DB15K and FB15K-YAGO15K; M, E, and C denote the margin, entropy, and CIS variants of ALMEA, respectively.}
\label{tab:cis_strategy}
\centering
\setlength{\tabcolsep}{2pt}
\renewcommand{\arraystretch}{1.02}
\resizebox{\columnwidth}{!}{%
\begin{tabular}{@{}clccccccccc@{}}
\toprule
\multirow{2}{*}{\textbf{Data}}
& \multirow{2}{*}{\textbf{Variants}}
& \multicolumn{3}{c}{\textbf{20\%}}
& \multicolumn{3}{c}{\textbf{50\%}}
& \multicolumn{3}{c}{\textbf{80\%}} \\
\cmidrule(lr){3-5}\cmidrule(lr){6-8}\cmidrule(l){9-11}
& & MRR & H@1& H@10
& MRR& H@1& H@10
& MRR& H@1& H@10\\
\midrule
\multirow{3}{*}{\rotatebox[origin=c]{90}{DB}}
& M & 55.4 & 46.5 & 72.2 & 73.3 & 66.4 & 85.4 & 82.4 & 77.5 & 90.5 \\
& E & 59.1 & 50.3 & 75.5 & 73.5 & 66.7 & 85.4 & 83.0 & 78.0 & 90.7 \\
& C & \textbf{59.4} & \textbf{50.8} & \textbf{75.6}
    & \textbf{74.2} & \textbf{67.4} & \textbf{86.2}
    & \textbf{83.8} & \textbf{79.1} & \textbf{92.4} \\
\midrule
\multirow{3}{*}{\rotatebox[origin=c]{90}{YG}}
& M & 44.3 & 35.9 & 60.4 & 66.0 & 58.6 & 79.4 & 77.0 & 71.0 & 87.3 \\
& E & 50.2 & 41.6 & 66.6 & 67.1 & 59.7 & 80.9 & 78.0 & 71.7 & 88.5 \\
& C & \textbf{50.9} & \textbf{42.1} & \textbf{67.7}
    & \textbf{68.8} & \textbf{61.3} & \textbf{82.5}
    & \textbf{78.3} & \textbf{72.3} & \textbf{89.0} \\
\bottomrule
\end{tabular}%
}
\end{table}

\par\noindent\textbf{Comparison with Alternative Selection Strategies. }
At 20\% supervision, two variants were introduced to validate the necessity of CIS. ALMEA$_{\text{w/ margin}}$ is the margin variant, which prioritizes the case with the smallest nearest-neighbor distance gap. In contrast, ALMEA$_{\text{w/ entropy}}$ is the entropy variant, which prioritizes the case with the highest binary entropy. As shown in Table ~\ref{tab:cis_strategy}, CIS performs best on both datasets and across all evaluation metrics, with performance that is on average 2.91\% higher than ALMEA$_{\text{w/ margin}}$ and 0.86\% higher than ALMEA$_{\text{w/ entropy}}$.

\par\noindent \textbf{Downstream Gains of CIS in Low-Resource Settings.} \label{sec:Downstream Gains of CIS}
To analyze how CIS translates candidate selection quality into alignment accuracy, we evaluate the framework under three resource-constrained scenarios: (i) \textbf{5\% Initial Supervision + 15\% Budget}, (ii) \textbf{10\% Initial Supervision + 10\% Budget}, and (iii) \textbf{15\% Initial Supervision + 5\% Budget}. Each setting utilizes a small set of randomly selected seed alignments (5%

As shown in Figure~\ref{fig:result of RQ2}, the \textbf{15\% Initial Supervision + 5\% Budget} configuration outperforms the empirical 20\% base supervision upper bound at both $\alpha = 0.1$ and $\alpha = 0.35$, yielding alignment improvements of 1.83\% and 2.87\%, respectively. This performance gain indicates that strategically selecting the final 5\% of training samples using CIS introduces significantly more valuable supervision than random sampling. However, under the \textbf{5\% Initial Supervision + 15\% Budget} setting, we observe that the alignment gains decrease because starting with a very small initial set (5\%) degrades the early-stage latent space alignment, which can cause the iterative selection algorithm to query false positive pairs. This demonstrates that maintaining a balance between the initial base supervision and iterative query budget is crucial for sample-efficient active MMEA. Detailed validation studies are provided in Supplementary Material~\ref{sec:Selection Strategy Analysis of CIS}, \ref{Iterative Learning without Control}, and \ref{Robustness of CIS}.

%% file: 06-Supplementary.tex
\section{Framework Details} \label{sec:derivation}

\subsection{Derivation Details}\label{sec:derivation details}
Table~\ref{tab:MMKG notation} summarizes the notations used throughout this paper, covering the multimodal knowledge graph definition, modal embedding construction, latent semantic modeling, CIS, and evaluation metrics.

\subsection{Additional Losses Details}\label{sec:Additional Losses Details}
For completeness, we provide additional explanations of the two auxiliary objectives used by ALMEA. Their mathematical definitions are given in the first and second lines of Eq.~\eqref{eq:vae_losses}, respectively, avoiding duplicate formulas and equation labels.

\noindent\textbf{Latent Distribution Normalization.}
The distribution matching loss (DML), defined in the first line of Eq.~\eqref{eq:vae_losses}, regularizes each modality-specific latent distribution toward a standard Gaussian distribution. This constraint structures and smooths the latent space while limiting irregular shifts in the learned means and variances.

\noindent\textbf{Missing Modality Reconstruction.}
CMI generates the missing latent variable $\tilde{z}^{m}$ through the reparameterization process and maps it to the calibrated latent embedding $\tilde{\mathbf{z}}^{m}$. The decoder then reconstructs the modality embedding $\hat{\mathbf{h}}^{m}$ to approximate the original embedding $\mathbf{h}^{m}$. The unimodal reconstruction loss (URL), defined in the second line of Eq.~\eqref{eq:vae_losses}, trains this reconstruction process for each modality $m\in\{g,r,a,v\}$.

\subsection{Optimization Formulation and Derivation}\label{sec:optimize formulation and derivation}
To learn efficiently from $Q_{\text{sorted}}$ under a limited budget, we formulate candidate selection as a coding-cost minimization problem. Let $K=|Q_{\text{sorted}}|$, and let $Z_{ij}=1$ indicate that candidate pair $q_i$ represents candidate pair $q_j$. The pairwise encoding cost is denoted by $A_{ij}$. The discrete formulation is
\begin{equation}
    \begin{aligned}
    \min_Z\quad
    &\lambda\|DZ\|_{1,p}
    +\sum_{j=1}^{K}\sum_{i=1}^{K}A_{ij}Z_{ij},\\
    \text{s.t.}\quad
    &\sum_{i=1}^{K}Z_{ij}=1,\quad \forall j,\\
    &Z_{ij}\in\{0,1\},\quad \forall i,j.
    \end{aligned}
\end{equation}
The mixed norm promotes the selection of a small number of representative rows, while the second term minimizes the total encoding cost. The assignment constraint ensures that each candidate pair is represented by one selected candidate.

\noindent\textbf{Relaxation.}
Because the binary formulation is combinatorial, we relax $Z_{ij}$ to the interval $[0,1]$ and set $p=2$. The resulting $\ell_{2,1}$-regularized problem is convex. The matrix $Z\in\mathbb{R}^{K\times K}$ contains the relaxed assignments, while $A\in\mathbb{R}^{K\times K}$ stores pairwise candidate dissimilarities. The vector $\mathbf{d}=[d_1,\ldots,d_K]^T\in\mathbb{R}^{K}$ contains the candidate-level diversity weights, and $D=\operatorname{diag}(\mathbf{d})\in\mathbb{R}^{K\times K}$ is the corresponding diagonal matrix. As defined in Eq.~\ref{eq:original_optimise_problem}, CIS solves
\begin{equation}
    \min_Z\ \lambda\|DZ\|_{2,1}+\langle A,Z\rangle
    \quad \text{s.t.}\quad
    \mathbf{1}^{T}Z=\mathbf{1}^{T},\quad Z\geq 0,
\end{equation}
where $\langle A,Z\rangle=\operatorname{tr}(A^{T}Z)$ and $\lambda$ controls row sparsity. The equality constraint ensures that every column of $Z$ sums to one, and the nonnegativity constraint gives valid relaxed assignments.

\noindent\textbf{Augmented Lagrangian Formulation. }
To separate the nonsmooth sparsity term from the simplex constraint, we introduce an auxiliary variable $P\in\mathbb{R}^{K\times K}$:
\begin{equation}
    \begin{aligned}
    \min_{Z,P}\quad &\lambda\|DZ\|_{2,1}+\langle A,P\rangle,\\
    \text{s.t.}\quad &Z=P,\quad \mathbf{1}^{T}P=\mathbf{1}^{T},\quad P\geq 0.
    \end{aligned}
    \label{eq:P}
\end{equation}
The corresponding augmented Lagrangian is
\begin{equation}
    \mathcal{L}_{\rho}(Z,P,\mu)
    =\lambda\|DZ\|_{2,1}+\langle A,P\rangle
    +\langle\mu,Z-P\rangle
    +\frac{\rho}{2}\|Z-P\|_{F}^{2},
\end{equation}
where $\mu\in\mathbb{R}^{K\times K}$ is the dual variable and $\rho>0$ is the ADMM penalty parameter. The sparsity coefficient is scaled as
\begin{equation}
    \lambda=\alpha\frac{\|A\|_{F}}{\sqrt{K}},
    \qquad \alpha\in[0.01,1],
    \label{sparsification_factor}
\end{equation}
so that larger values of $\alpha$ impose stronger row sparsity.

\noindent\textbf{Derivation Details. }
For fixed $P^{(t)}$ and $\mu^{(t)}$, the terms associated with the $Z$-update can be expanded by completing the square:
\begin{equation}
    \begin{aligned}
    &\left\langle\mu^{(t)},Z-P^{(t)}\right\rangle
    +\frac{\rho}{2}\left\|Z-P^{(t)}\right\|_F^2 \\
    &=\frac{\rho}{2}\left(
    \left\|Z-P^{(t)}\right\|_F^2
    +2\left\langle\frac{\mu^{(t)}}{\rho},Z-P^{(t)}\right\rangle
    +\frac{\left\|\mu^{(t)}\right\|_F^2}{\rho^2}
    \right)
    -\frac{\left\|\mu^{(t)}\right\|_F^2}{2\rho} \\
    &=\frac{\rho}{2}\left\|Z-P^{(t)}+\frac{\mu^{(t)}}{\rho}\right\|_F^2
    -\frac{\left\|\mu^{(t)}\right\|_F^2}{2\rho}.
    \end{aligned}
\end{equation}

For fixed $Z^{(t+1)}$ and $\mu^{(t)}$, the terms associated with the $P$-update are
\begin{equation}
    \begin{aligned}
    &\left\langle A,P\right\rangle
    +\left\langle\mu^{(t)},Z^{(t+1)}-P\right\rangle
    +\frac{\rho}{2}\left\|Z^{(t+1)}-P\right\|_F^2 \\
    &=\frac{\rho}{2}\left\|P-Z^{(t+1)}\right\|_F^2
    +\left\langle A-\mu^{(t)},P\right\rangle+\mathrm{const} \\
    &=\frac{\rho}{2}\left\|P-\left(
    Z^{(t+1)}+\frac{\mu^{(t)}-A}{\rho}
    \right)\right\|_F^2+\mathrm{const}.
    \end{aligned}
\end{equation}

\subsection{Final Update Formulas}\label{sec:final}

\noindent\textbf{Update of $Z$.}
\begin{equation}
    \begin{aligned}
    Z^{(t+1)}
    &=\operatorname*{arg\,min}_{Z}
    \lambda\|DZ\|_{2,1}
    +\left\langle\mu^{(t)},Z-P^{(t)}\right\rangle
    +\frac{\rho}{2}\left\|Z-P^{(t)}\right\|_F^2 \\
    &=\operatorname*{arg\,min}_{Z}
    \lambda\|DZ\|_{2,1}
    +\frac{\rho}{2}\left\|Z-P^{(t)}+\frac{\mu^{(t)}}{\rho}\right\|_F^2 \\
    &=\operatorname*{arg\,min}_{Z}
    \frac{\lambda}{\rho}\|DZ\|_{2,1}
    +\frac{1}{2}\left\|Z-P^{(t)}+\frac{\mu^{(t)}}{\rho}\right\|_F^2.
    \end{aligned}
    \label{update Z}
\end{equation}
This is a weighted row-sparsity proximal step with $P^{(t)}$ and $\mu^{(t)}$ fixed.

\noindent\textbf{Update of $P$.}
\begin{equation}
    \begin{aligned}
    P^{(t+1)}
    &=\operatorname*{arg\,min}_{P\in\Delta}
    \left\langle A,P\right\rangle
    +\left\langle\mu^{(t)},Z^{(t+1)}-P\right\rangle
    +\frac{\rho}{2}\left\|Z^{(t+1)}-P\right\|_F^2 \\
    &=\operatorname*{arg\,min}_{P\in\Delta}
    \left\|P-\left(Z^{(t+1)}+\frac{\mu^{(t)}-A}{\rho}\right)\right\|_F^2 \\
    &=\Pi_{\Delta}\left(Z^{(t+1)}+\frac{\mu^{(t)}-A}{\rho}\right),
    \end{aligned}
    \label{update P}
\end{equation}
where $\Pi_{\Delta}$ denotes projection onto
$\Delta=\{P\mid\mathbf{1}^{T}P=\mathbf{1}^{T},\ P\geq0\}$.

\noindent\textbf{Update of $\mu$.}
\begin{equation}
    \mu^{(t+1)}=\mu^{(t)}+\rho\left(Z^{(t+1)}-P^{(t+1)}\right).
    \label{update mu}
\end{equation}
The iterations terminate when the primal residual satisfies
$\|Z^{(t+1)}-P^{(t+1)}\|_F\leq\epsilon$ or when $T_{\mathrm{iter}}$ iterations have been completed.

\subsection{Detailed Derivation of Time Complexity}\label{sec:derivation of time complexity}
We analyze the computational cost of CIS. Computing the source-to-target distance matrix for $M$ source entities and $N$ target entities requires $O(MNd)$ operations, where $d$ is the embedding dimension. Extracting the unlabeled submatrix and identifying mutual nearest neighbors requires $O(M^U N^U)$. The total cost of constructing the candidate pool in Eq.~\ref{eq:Q} is therefore
\begin{equation}
    \mathcal{C}_{Q}
    =O\!\left(MNd+M^U N^U\right).
\end{equation}

Constructing the dissimilarity matrix $A$ in Eq.~\ref{eq:CIS_metrics} requires $O(K^2d)$ operations, where $K=|Q_{\mathrm{sorted}}|$. Computing the diversity matrix $C$ for $K$ candidates and $|L|$ labeled pairs requires $O(K|L|d)$ operations. Reducing $C$ to the diversity weights in the same equation requires an additional $O(K|L|)$ operations. Thus, feature construction costs

\begin{equation}
    O\!\left(K^2d+K|L|d+K|L|\right).
\end{equation}

The $Z$, $P$, and dual-variable updates each operate on $K\times K$ matrices. One ADMM iteration therefore costs $O(K^2)$, and $T_{\mathrm{iter}}$ iterations cost
\begin{equation}
    O\!\left(T_{\mathrm{iter}} K^2\right).
\end{equation}
Combining these components gives
\begin{equation}
    O\!\left(MNd+M^U N^U+K^2d+K|L|d+K|L|+T_{\mathrm{iter}}K^2\right).
\end{equation}

Since the candidate pool is obtained by filtering the global entity space, $K$ is typically much smaller than $M$ and $N$. It can be reduced further by retaining only candidate pairs whose confidence exceeds a specified threshold. Therefore, for the balanced setting $M\approx N$, $K$ can be treated as a small user-controlled quantity relative to the graph size. The number of ADMM iterations $T_{\mathrm{iter}}$ is also fixed independently of the entity scale.

Using $M=M^L+M^U$ and $M^L,M^U\leq M$, the terms involving the labeled and unlabeled subsets are upper-bounded by
\begin{equation}
    O\!\left(M^2d+Md+M+M^2\right).
\end{equation}
Thus, under the balanced-graph assumption, the worst-case complexity is dominated by $O(M^2d)$. For strongly imbalanced source and target KGs, the unsimplified expression above should be used instead.

\section{Additional Experimental Details}\label{sec:more exp details}

\subsection{Benchmark Dataset Details}\label{ap:data} 
We evaluate the MMEA task on two widely recognized benchmark MMKGs: FB15K-DB15K and FB15K-YAGO15K~\cite{OoroRubio2019AnsweringVQ, Garcia-DuranN18, liu2019mmkg}. Each MMKG includes relational \cite{liu2019mmkg}, visual \cite{OoroRubio2019AnsweringVQ}, textual, and numerical attributes \cite{Garcia-DuranN18}. Relations and Attributes represent unique relationship and attribute modality information within the complete dataset. Since an entity can have multiple attributes, attribute triples may appear multiple times, as do relationship triples.
Ground-truth (GT) refers to the true entity alignments, where each source entity is uniquely paired with a target entity. Note that some source entities may lack image modality or a corresponding target entity.
Entities lacking image information are pre-processed through random sampling using a normal distribution to ensure consistent input for the visual modality. The overall dataset statistics are summarized in Table~\ref{Statist_datasets}.

\begin{table}[ht]
  \caption{Dataset Summary. $|GT|$ represents the ground-truth entity alignments, indicating the number of aligned entity pairs between the source and target datasets. Specifically, $|GT|$(FB15K-DB15K) denotes the number of aligned entity pairs between FB15K and DB15K, while $|GT|$(FB15K-YAGO15K) denotes the aligned pairs between FB15K and YAGO15K. Some source entities may lack image modality or a corresponding target entity.}
  \label{Statist_datasets}
  \centering
  \footnotesize
  \setlength\tabcolsep{3pt}
  \resizebox{\textwidth}{!}{
  \begin{tabular}{lccrrrcrr}
    \toprule
    $\mathcal{G}$ & $|\mathcal{E}|$ & $|\mathcal{R}|$ & \#Triples($\mathcal{R}$) & $\mathcal{A}$ & \#Triples($\mathcal{A}$) & $|\mathcal{V}|$ & $|GT|$(FB15K-DB15K) & $|GT|$(FB15K-YAGO15K) \\
    \midrule
    FB15K & 14,951 & 1,345 & 592,213 & 116 & 29,385 & 13,444 & 12,846 & 11,199 \\
    DB15K & 12,842 & 279 & 89,197 & 225 & 48,080 & 12,837 & 12,846 & -- \\
    YAGO15K & 15,404 & 32 & 122,886 & 7 & 23,532 & 11,194 & -- & 11,199 \\
    \bottomrule
  \end{tabular}}
\end{table}

\subsection{Baseline Details}\label{ap:baseline}
For classical KG embedding methods, we consider the following approaches.
\begin{itemize}%
    \item \textbf{TransE}~\cite{bordes2013translating}: Learns entity and relation embeddings in a continuous space, aligning entities through vector addition.
    \item \textbf{GCN-Align}~\cite{wang2018cross}: Integrates structural and attribute information by mapping entities into a unified embedding space via a graph convolutional network.
    \item \textbf{SEA}~\cite{pei2019semi}: Enhances entity embeddings using degree-aware adjustments and optimizes alignment performance through iterative training with both labeled and unlabeled entity pairs.
\end{itemize}

For multimodal KG embedding methods, we consider the following approaches.
\begin{itemize}%
    \item \textbf{MMEA}~\cite{chen2020mmea}: Maps visual, attribute, and relational modalities into a unified low-dimensional embedding space, and fuses these modalities to improve alignment accuracy.
    \item \textbf{EVA}~\cite{liu2021visual}: Emphasizes visual similarity as a key alignment factor, leveraging visual, relational, and attribute modalities for joint representation and unsupervised learning.
    \item \textbf{MSNEA}~\cite{chen2022multi}: Utilizes visual modalities to enhance relational and attribute-based alignment, focusing on cross-modal interactions.
    \item \textbf{MCLEA}~\cite{lin2022multi}: Employs contrastive learning \cite{chen2020simple} to improve alignment by capturing intra-modal mutual information and cross-modal distribution differences, thereby strengthening modal fusion consistency.
    \item \textbf{GEEA}~\cite{guo2023revisit}: Uses VAE-based generative models to reconstruct embeddings across KGs, enhancing information sharing and feature associations to improve alignment.
    \item \textbf{MEAformer}~\cite{chen2023meaformer}: Introduces a multi-head attention mechanism~\cite{vaswani2017attention} to dynamically adjust modality weights, mitigating information loss during fusion and extracting core semantics for optimized alignment.
    \item \textbf{OTMEA}~\cite{wang2025otmea}: Mitigates semantic heterogeneity between modalities through optimal transport (OT)~\cite{cao2022otkge}, specifically by mapping visual, neighbourhood, attribute, and relational modalities into a unified distributional space. This approach aims to approximate the geometric structure between modalities, thereby learning more robust joint embeddings.
    \item \textbf{SimDiff}~\cite{li2024simdiff}: By mapping neighbourhood, attribute, relational, and visual modalities onto a unified latent space, perturbation modal embeddings are obtained through a Gaussian noise augmentation paradigm. Subsequently, a progressive denoising process implemented with stacked linear layers facilitates intermodal information transfer and thereby mitigates modal heterogeneity.
\end{itemize}

\subsection{Implementation Details} 
The number of training iterations, which varies depending on the dataset, is summarized in Table ~\ref{tab:hyperparams}. All experiments were conducted on a Tesla A100 GPU.
For models that could not be executed locally, including TransE, GCN-Align, SEA, MMEA, and EVA, we refer to the reported results from~\cite{chen2020mmea}. The results for MSNEA, MCLEA, GEEA, MEAformer, OTMEA, and SimDiff were obtained from our local experimental runs using default parameter settings. In our experiments, $\lambda_{SCL}$, $\lambda_{DML}$, $\lambda_{URL}$, and $\lambda_{CMAL}$ are all set to 1.
The VAE module comprises a unimodal encoder, a hidden layer, and a unimodal decoder, each with dimensionality 300, and a joint modal decoder also of dimensionality 300. The learning rate is fixed at 0.001 with Adam optimization, a batch size of 3500, and a constant scheduler. 

\begin{table}[t]
  \caption{Hyperparameter settings in the main experiments. Training Configuration Settings}
  \label{tab:hyperparams}
  \centering
  \small
  \setlength\tabcolsep{4pt}
  \resizebox{\textwidth}{!}{
  \begin{tabular}{lccccccccc}
    \toprule
    \textbf{Datasets} & $T_{\mathrm{total}}$ & $T_{\mathrm{warm}}$ & $T_{\mathrm{inter}}$ & $T_r$ & $T_{\mathrm{iter}}$ & \textbf{batch-size} & \makecell{\textbf{masked dropping} \\ \textbf{rate}} & \textbf{learning rate} & \makecell{\textbf{sparsification factor $\alpha$} \\ (20\%, 50\%, 80\%)}\\
    \midrule
    FB15K-DB15K & 750 & 500 & 50 & 5 & 1000 & 3,500 & 0.45 & 0.001 & [0.35, 0.45, 0.95]\\
    FB15K-YAGO15K & 550 & 300 & 50 & 5 & 1000 & 3,500 & 0.50 & 0.001 & [0.25, 0.35, 0.45]\\
    \bottomrule
  \end{tabular}}
\end{table}

\subsection{Auxiliary-Loss Ablation}\label{sec:auxiliary_loss_ablation}
Table~\ref{tab:auxiliary_loss_ablation} reports the loss-level ablations corresponding to the auxiliary objectives described above. Each auxiliary objective contributes to performance, although its effect is smaller than that of the core semantic calibration loss reported in the main manuscript.

\begin{table}[t]
    \centering
    \caption{Auxiliary-loss ablation under 80\% supervision. All metrics are reported in \%.}
    \label{tab:auxiliary_loss_ablation}
    \setlength{\tabcolsep}{4pt}
    \begin{tabular}{lcccccc}
        \toprule
        & \multicolumn{3}{c}{FB15K-DB15K}
        & \multicolumn{3}{c}{FB15K-YAGO15K} \\
        \cmidrule(lr){2-4}\cmidrule(lr){5-7}
        Variant & MRR & Hits@1 & Hits@10 & MRR & Hits@1 & Hits@10 \\
        \midrule
        ALMEA$_{\text{w/o CIS}}$ & \textbf{84.83} & \textbf{80.45} & \textbf{93.28} & \textbf{80.10} & \textbf{74.48} & \textbf{90.16} \\
        w/o $\mathcal{L}_{DML}$ & 82.96 & 77.80 & 91.36 & 78.33 & 72.30 & 88.86 \\
        w/o $\mathcal{L}_{URL}$ & 83.46 & 78.50 & 91.51 & 78.56 & 72.67 & 88.99 \\
        w/o $\mathcal{L}_{CMAL}$ & 83.43 & 78.25 & 91.38 & 78.33 & 72.45 & 88.72 \\
        \bottomrule
    \end{tabular}
\end{table}

\subsection{Statistical Significance Test}\label{Statistical Significance Test}
\begin{table}[t]
\centering
\caption{Summary of statistical significance (paired t-test over 5 runs). Significance levels: * = $p<0.05$, ** = $p<0.01$, *** = $p<0.001$, --- = not significant.}
\label{tab:significance_summary}
\setlength{\tabcolsep}{5pt}
\resizebox{1\textwidth}{!}{\begin{tabular}{lcccccc}
\toprule
\textbf{Dataset \& Ratio} & \textbf{MCLEA} & \textbf{MSNEA} & \textbf{GEEA} & \textbf{MEAformer} & \textbf{OTMEA} & \textbf{SimDiff (SoTA)} \\
\midrule
FB15K-DB15K (20\%)  & *** & *** & **  & *** & **  & **  \\
FB15K-DB15K (50\%)  & *** & **  & *** & *** & **  & *** \\
FB15K-DB15K (80\%)  & *** & *   & *   & *** & --- & *** \\
FB15K-YAGO15K (20\%) & *** & *** & *** & *** & **  & *** \\
FB15K-YAGO15K (50\%) & *** & **  & *** & *** & --- & *** \\
FB15K-YAGO15K (80\%) & *** & *   & *** & *** & --- & *** \\
\bottomrule
\end{tabular}}
\end{table}
To further evaluate the efficacy of \textbf{ALMEA}, we conducted paired t-tests~\cite{student1908probable,demvsar2006statistical} against all benchmark models. As shown in Table~\ref{tab:significance_summary}, the improvements achieved by ALMEA are statistically significant in most cases. Specifically, across all datasets and training configurations, ALMEA reached significance at \(p < 0.01\) in the majority of comparisons, indicating that its alignment gains are statistically reliable rather than due to random fluctuations. Cases with \(p > 0.05\) occurred only when compared with OTMEA, which exhibits substantially greater variation in alignment accuracy than other benchmarks. For example, under the MRR, Hits@1, and Hits@10 metrics, MSNEA shows an average standard deviation of \(\pm 5.97\%\), OTMEA \(\pm 4.58\%\), whereas ALMEA maintains a low deviation of only \(\pm 0.47\%\). Such pronounced fluctuations make it difficult to establish statistical significance under repeated experimental settings.

\subsection{Quality of Synthesized Modality}\label{Quality of Synthesized Modality}
We find that reconstructing synthetic modalities reflects the model’s ability to impute missing low-level details, while evaluating these reconstructions within a specific semantic domain further suggests its capacity for high-level semantic generalisation. So, to evaluate the quality of synthesized modalities, we use 20\% of the training set and compare ALMEA with GEEA, which employs a VAE for synthetic modality generation.
Table~\ref{RQ4:bulgaria},\ref{RQ4:mayor},\ref{RQ4:river thames} shows that both ALMEA and GEEA effectively reconstruct visual embeddings, but ALMEA outperforms GEEA in reconstructing neighbors and attributes. ALMEA demonstrates high-level semantic generalisation, particularly in the Attribute category. For example, in Table~\ref{RQ4:bulgaria}, ALMEA can preserve higher-level semantic relationships; it not only identifies “populationDensity” but also identifies three semantically similar candidates. As for Table~\ref{RQ4:river thames}, ALMEA identifies a greater number of semantically valuable candidates. Furthermore, the reconstruction results for Neighbourhood demonstrate that ALMEA can mitigate overvaluation. Specifically, ALMEA can reconstruct rich modality details, helping mitigate the introduction of local modality noise and thereby reducing the impact of overvaluation.

\begin{table*}[ht]
    \caption{Qualitative comparison of imputed modalities for Bulgaria on FB15K. For each imputed modality, the nearest neighbors are retrieved in the embedding space. Blue text denotes retrieved items that exactly match the ground truth, while underlined text denotes additional non-ground-truth neighbors returned by KNN.}
    \centering
    \centering
    \small
    \setlength\tabcolsep{4pt}
    \resizebox{\textwidth}{!}{
    \begin{tabular}{
        >{\raggedright\arraybackslash}p{0.33\linewidth}             
        >{\raggedright\arraybackslash}p{0.33\linewidth}             
        >{\raggedright\arraybackslash}p{0.33\linewidth}}
        \toprule
        \textbf{Ground Truth Nearest Neighbor} & \textbf{ALMEA Nearest Neighbor} & \textbf{GEEA Nearest Neighbor} \\
        \midrule
        
        \textbf{\# Entity Name}: Bulgaria & \textbf{\# Entity Name}: Bulgaria & \textbf{\# Entity Name}: Bulgaria \\
        \includegraphics[width=0.5\linewidth]{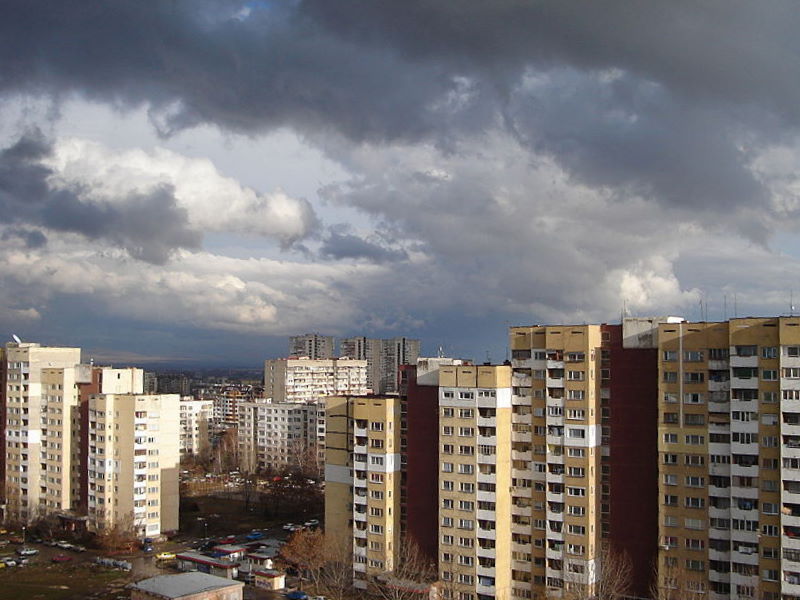} & 
        \includegraphics[width=0.5\linewidth]{figure/Bulgaria.jpg} & 
        \includegraphics[width=0.5\linewidth]{figure/Bulgaria.jpg} \\
        
        \textbf{\# Neighborhood}: Belgium, Hungary, Greece, Romania, Croatia & 
        \textbf{\# Neighborhood}: \textcolor{blue}{Hungary}, \textcolor{blue}{Greece}, \textcolor{blue}{Romania}, \textcolor{blue}{Croatia}, \underline{Slovenia}, \underline{Chile}, \underline{Colombia}, \underline{Lithuania}, \underline{Taiwan}, \underline{Iceland} & 
        \textbf{\# Neighborhood}: \textcolor{blue}{Belgium}, \textcolor{blue}{Greece}, \underline{Slovenia}, \underline{Chile}, \underline{Lithuania}, \underline{Iceland} \\
        
        \textbf{\# Attribute}: iso\_numeric, longitude, latitude, area, date\_founded, population\_number, calling\_code & 
        \textbf{\# Attribute}: \textcolor{blue}{iso\_numeric}, \textcolor{blue}{longitude}, \textcolor{blue}{latitude}, \textcolor{blue}{area}, \textcolor{blue}{date\_founded}, \textcolor{blue}{population\_number}, \textcolor{blue}{calling\_code}, \underline{date\_founded}, \underline{date\_dissolved}, \underline{wasCreatedOnDate}, \underline{populationDensity} & 
        \textbf{\# Attribute}: \textcolor{blue}{iso\_numeric}, \textcolor{blue}{longitude}, \textcolor{blue}{latitude}, \textcolor{blue}{area}, \textcolor{blue}{date\_founded}, \textcolor{blue}{population\_number}, \textcolor{blue}{calling\_code}, \underline{populationDensity} \\ 
        \bottomrule
    \end{tabular}
    }
    \label{RQ4:bulgaria}
\end{table*}
\begin{table*}[ht]
    \caption{Qualitative comparison of imputed modalities for Mayor on FB15K. For each imputed modality, the nearest neighbors are retrieved in the embedding space. Blue text denotes retrieved items that exactly match the ground truth, while underlined text denotes additional non-ground-truth neighbors returned by KNN.}
    \centering
    \small
    \setlength\tabcolsep{4pt}
    \resizebox{\textwidth}{!}{
    \begin{tabular}{
        >{\raggedright\arraybackslash}p{0.33\linewidth}
        >{\raggedright\arraybackslash}p{0.33\linewidth}
        >{\raggedright\arraybackslash}p{0.33\linewidth}}
        \toprule
        \textbf{Ground Truth Nearest Neighbor} & \textbf{ALMEA Nearest Neighbor} & \textbf{GEEA Nearest Neighbor} \\
        \midrule
        
        \textbf{\# Entity Name}: Mayor & \textbf{\# Entity Name}: Mayor & \textbf{\# Entity Name}: Mayor \\
        \includegraphics[width=0.5\linewidth]{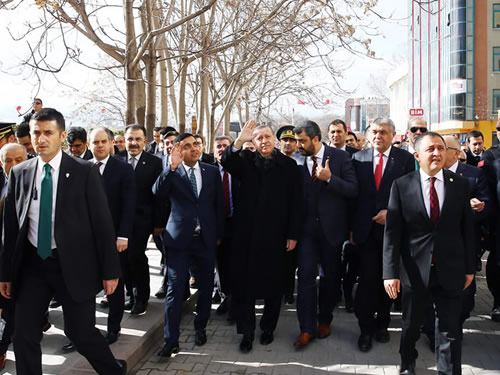} &
        \includegraphics[width=0.5\linewidth]{figure/mayor.jpg} &
        \includegraphics[width=0.5\linewidth]{figure/mayor.jpg} \\
        
        \textbf{\# Neighborhood}: Lafayette\_Louisiana, Dnipropetrovsk, Laredo\_Texas &
        \textbf{\# Neighborhood}: \textcolor{blue}{Lafayette\_Louisiana}, \textcolor{blue}{Dnipropetrovsk}, \textcolor{blue}{Laredo\_Texas}, \underline{Municipal Council} &
        \textbf{\# Neighborhood}: \textcolor{blue}{Lafayette\_Louisiana}, \textcolor{blue}{Laredo\_Texas}, \underline{California}, \underline{Municipal\_Councils}, \underline{United\_States} \\
        
        \textbf{\# Attribute}: Null &
        \textbf{\# Attribute}: \underline{date\_established}, \underline{date\_founded}, \underline{date\_of\_birth}, \underline{height\_meters}, \underline{initial\_release\_date}, \underline{interest} &
        \textbf{\# Attribute}: \underline{activeYearsEndYear}, \underline{birthYear}, \underline{activeYearsStartYear}, \underline{birthDate} \\
        
        \bottomrule
    \end{tabular}
    }
    \label{RQ4:mayor}
\end{table*}
\begin{table*}[ht]
    \caption{Qualitative comparison of imputed modalities for River Thames on FB15K. For each imputed modality, the nearest neighbors are retrieved in the embedding space. Blue text denotes retrieved items that exactly match the ground truth, while underlined text denotes additional non-ground-truth neighbors returned by KNN.}
    \centering
    \small
    \setlength\tabcolsep{4pt}
    \resizebox{\textwidth}{!}{
    \begin{tabular}{
        >{\raggedright\arraybackslash}p{0.33\linewidth}
        >{\raggedright\arraybackslash}p{0.33\linewidth}
        >{\raggedright\arraybackslash}p{0.33\linewidth}}
        \toprule
        \textbf{Ground Truth Nearest Neighbor} & \textbf{ALMEA Nearest Neighbor} & \textbf{GEEA Nearest Neighbor} \\
        \midrule
        
        \textbf{\# Entity Name}: River Thames & \textbf{\# Entity Name}: River Thames & \textbf{\# Entity Name}: River Thames \\
        \includegraphics[width=0.5\linewidth]{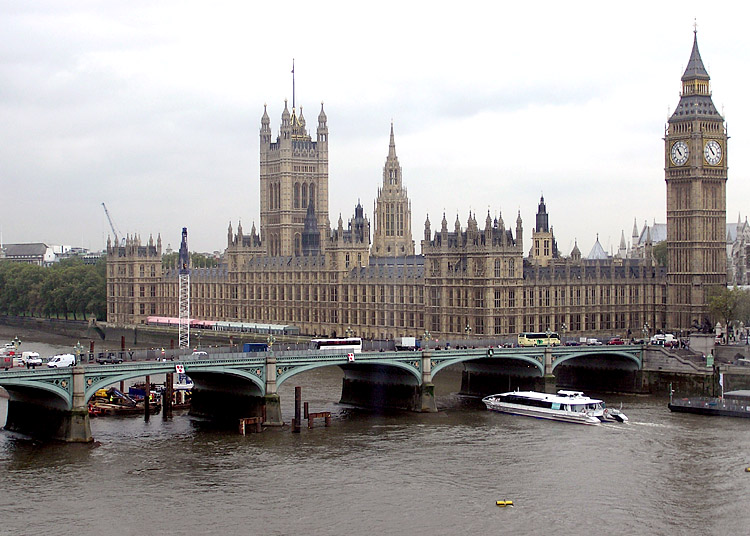} &
        \includegraphics[width=0.5\linewidth]{figure/River_Thames.jpg} &
        \includegraphics[width=0.5\linewidth]{figure/River_Thames.jpg} \\
        
        \textbf{\# Neighborhood}: London\_Borough\_of\_Newham, London\_Borough\_of\_Tower\_Hamlets, City\_of\_London, Richmond\_London &
        \textbf{\# Neighborhood}: \textcolor{blue}{London\_Borough\_of\_Newham}, \textcolor{blue}{London\_Borough\_of\_Tower\_Hamlets}, \textcolor{blue}{City\_of\_London}, \textcolor{blue}{Richmond\_London}, \underline{Naveen\_Andrews}, \underline{United\_Kingdom}, \underline{Greater\_London}, \underline{Lincoln\_England}, \underline{St\_Helens\_Merseyside}, \underline{London\_Borough\_of\_Hackney} &
        \textbf{\# Neighborhood}: \textcolor{blue}{London\_Borough\_of\_Newham}, \textcolor{blue}{London\_Borough\_of\_Tower\_Hamlets}, \textcolor{blue}{City\_of\_London}, \underline{City\_of\_Westminster}, \underline{Greater\_London}, \underline{London\_Borough\_of\_Hackney} \\
        
        \textbf{\# Attribute}: discharge, length, source\_elevation, basin\_area &
        \textbf{\# Attribute}: \textcolor{blue}{discharge}, \textcolor{blue}{length}, \textcolor{blue}{source\_elevation}, \textcolor{blue}{basin\_area}, \underline{date\_founded}, \underline{latitude}, \underline{longitude}, \underline{elevation}, \underline{areaCode} &
        \textbf{\# Attribute}: \textcolor{blue}{discharge}, \textcolor{blue}{length}, \textcolor{blue}{source\_elevation}, \textcolor{blue}{basin\_area}\\
        
        \bottomrule
    \end{tabular}
    }
    \label{RQ4:river thames}
\end{table*}

\subsection{Mitigating CI under Limited Supervision}\label{supp:recovered_case_analysis}
\begin{figure*}[t]
  \centering
  \includegraphics[width=\textwidth]{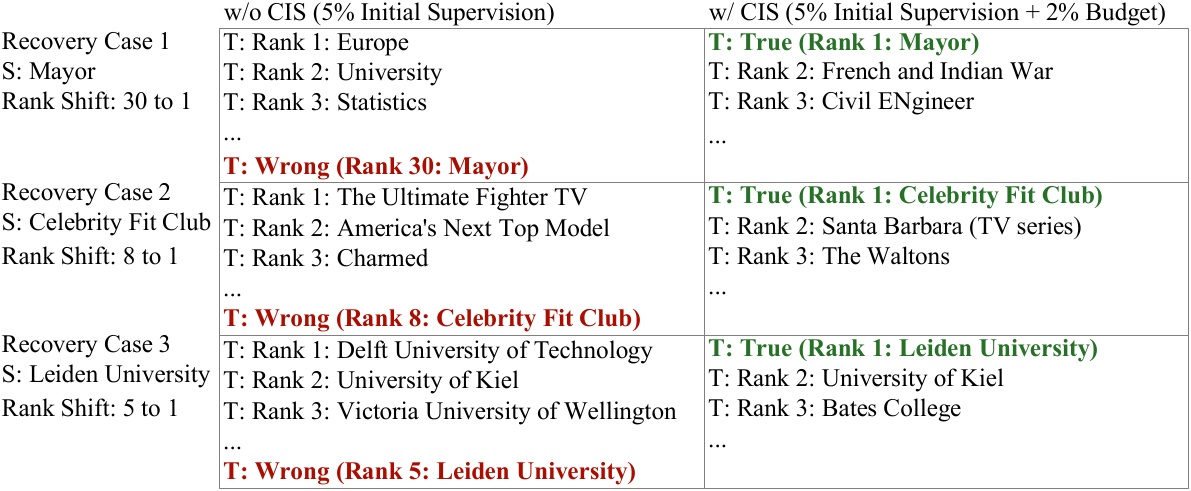}
  \caption{Random recovery case under Limited Supervision. With only 5\% of the supervision and just 2\% of the budget, CIS elevated GT's ranking to top 1.}
  \label{fig:fig3_CIS}
\end{figure*}

In our experiments, we found that the calibration paradigm based on the VAE's latent space requires richer supervisory signals to mitigate suppression phenomena. Consequently, we found that in low-resource scenarios, the benefits of LSC can be limited by insufficient supervision. Therefore, we improve upon Iterative Learning \cite{sun2018bootstrapping} and propose Controlled Iterative Selection (CIS) to enhance LSC's effectiveness in resource-constrained scenarios. This controllable paradigm efficiently labels candidate pairs most beneficial to the current supervision environment with minimal budget. As illustrated in Figure \ref{fig:fig3_CIS}, we randomly selected three recovery cases. To demonstrate CIS's efficacy, we employed only 5\% of the training set whilst alleviating LSC's effectiveness in low-resource scenarios with a mere 2\% budget. The rank orderings across all three cases were corrected from top x to top 1. In the next section, we provide core details of ALMEA to demonstrate how CMI, LSC, and CIS operate.

\subsection{CIS Selection Strategy Analysis under Limited Supervision} \label{sec:Selection Strategy Analysis of CIS}
\begin{table*}[ht]
  \centering
  \setlength\tabcolsep{0.95pt}
  \caption{CIS Selection Strategy Analysis under Limited Supervision. Results are under 20\%, 50\%, and 80\% valid alignment pairs settings.
  }
  \label{Selection Strategy Analysis of CIS}
  \resizebox{1\textwidth}{!}{
    \begin{tabular}{lccccccccc}
        \toprule
        & \multicolumn{3}{c}{FB15K-DB15K (20\%)} & \multicolumn{3}{c}{FB15K-DB15K (50\%)} & \multicolumn{3}{c}{FB15K-DB15K (80\%)} \\
        \cmidrule(r){2-4} \cmidrule(lr){5-7} \cmidrule(l){8-10}
        Methods & MRR(\%) & Hits@1(\%) & Hits@10(\%) & MRR(\%) & Hits@1(\%) & Hits@10(\%) & MRR(\%) & Hits@1(\%) & Hits@10(\%) \\
        \midrule

        ALMEA$_{\text{w/ margin}}$ & 55.42 & 46.52 & 72.24 & 73.30 & 66.40 & 85.39 & 82.40 & 77.53 & 90.54\\

        ALMEA$_{\text{w/ entropy}}$ & 59.09 & 50.27 & 75.45 & 73.52 & 66.74 & 85.40 & 82.95 & 77.98 & 90.74\\

        ALMEA$_{\text{w/ CIS}}$ & \textbf{59.43} & \textbf{50.79} & \textbf{75.64} & \textbf{74.16} & \textbf{67.40} & \textbf{86.16} & \textbf{83.80} & \textbf{79.08} & \textbf{92.41}\\
    \end{tabular}}
    
    \resizebox{1\textwidth}{!}{
    \begin{tabular}{lccccccccc}
        \toprule
        & \multicolumn{3}{c}{FB15K-YAGO15K (20\%)} & \multicolumn{3}{c}{FB15K-YAGO15K (50\%)} & \multicolumn{3}{c}{FB15K-YAGO15K (80\%)} \\
        \cmidrule(r){2-4} \cmidrule(lr){5-7} \cmidrule(l){8-10}
        Methods & MRR(\%) & Hits@1(\%) & Hits@10(\%) & MRR(\%) & Hits@1(\%) & Hits@10(\%) & MRR(\%) & Hits@1(\%) & Hits@10(\%) \\
        \midrule

        ALMEA$_{\text{w/ margin}}$ & 44.30 & 35.89 & 60.43 & 66.01 & 58.61 & 79.40 & 77.00 & 70.96 & 87.32\\

        ALMEA$_{\text{w/ entropy}}$ & 50.20 & 41.55 & 66.55 & 67.10 & 59.67 & 80.90 & 78.01 & 71.74 & 88.45\\

        ALMEA$_{\text{w/ CIS}}$ & \textbf{50.92} & \textbf{42.11} & \textbf{67.70} & \textbf{68.77} & \textbf{61.33} & \textbf{82.50} & \textbf{78.34} & \textbf{72.27} & \textbf{88.95}\\
        \bottomrule
    \end{tabular}}
\end{table*}
In active learning, sample selection strategies are typically categorised as uncertainty-based, representativeness-based, or diversity-based \cite{settles2009active}. Although iterative learning is at the core of our approach rather than active learning, CIS also relies on sample labelling. Consequently, within the paradigm of sample labelling, the CIS design primarily combines the principles of representativeness and diversity. To investigate whether uncertainty-style signals can further enhance the performance of MMEA in this iterative setting, we constructed the following two uncertainty-based variants of CIS.

\noindent\textbf{Margin Uncertainty-based Method.}
Based on Eq.~\ref{eq:Q}, the core objective of CIS is to obtain $Q_{\text{sorted}}$. Thus, we define
\[
    d_i(j) = \left\| \mathbf{h}_{S,i} - \mathbf{h}_{T,j} \right\|^2
\]
, where \( j \in 1, \ldots, N \) enumerates all target entities and \( i \in 1, \ldots, M \) enumerates all source entities. And then let the nearest and second-nearest targets be:
\[
    j_1 = \arg \min_j d_i(j), \quad j_2 = \arg \min_{j \neq j_1} d_i(j)
\]
So, we have distance-based margin as \( Q_m(i) = d_i(j_2) - d_i(j_1) \). A smaller margin indicates higher alignment uncertainty, so we select candidates with the smallest margins~\cite{settles2009active}. We denote this paradigm as ALMEA$_{\text{w/ margin}}$.

\noindent\textbf{Pseudo-entropy Uncertainty-based Method.}
Since no probability distribution was used during the construction of \( Q \), we adopt the entropy of the binomial Bernoulli distribution~\cite{phoenix1992elements} to establish a pseudo-entropy uncertainty-based paradigm. We start from the distance, compute the nearest target distance as:
\[
    Q_e(i) = -(s_i \log s_i + (1 - s_i) \log(1 - s_i))
\]
So, we have a larger \( Q_{\text{entropy}}(i) \), which signifies more hesitant and uncertain alignment decision. We have labelled this method as ALMEA$_{\text{w/ entropy}}$.

As shown in the Table~\ref{Selection Strategy Analysis of CIS}. We found that margin-based uncertainty behaved unexpectedly, as algorithm selects the most uncertain candidate pairs, such as $(i,j)$, where the model cannot determine whether its prediction for $i$ corresponds to $j_1$ or $j_2$. Therefore, expert annotations are required to resolve the model's prediction ambiguity. This approach results in suboptimal performance of $Q_m$.

And, the entropy-based method outperforms the margin-based method, but still performs weaker than the CIS method. We found that for the MMEA task, the entropy paradigm selects samples where the model is most hesitant, rather than those it finds most difficult to distinguish and least confident (as chosen by the margin-based method). Therefore, the entropy method significantly outperforms the margin-based approach. Although the difference is marginal, we observed during our experiments that the entropy method’s selection of samples from the ``hesitation region'' inevitably introduces noise, further confirming its slightly inferior performance relative to the CIS method.

\subsection{CIS under Unlimited Budget (Iterative Learning without Control)} \label{Iterative Learning without Control}
\begin{table*}[ht]
  \centering
  \setlength\tabcolsep{0.95pt}
  \caption{CIS under unlimited budget, also be described as iterative learning (IL). Here, ALMEA$_{\text{w/ CIS}}$ stands for ALMEA with Controlled Iterative Selection (using a limited budget). ALMEA$_{\text{w/ IL}}$ denotes ALMEA with Iterative Learning (use an unlimited budget).
  }
  \label{CIS under Unlimited Budget}
  \resizebox{1\textwidth}{!}{
    \begin{tabular}{lccccccccc}
        \toprule
        & \multicolumn{3}{c}{FB15K-DB15K (20\%)} & \multicolumn{3}{c}{FB15K-DB15K (50\%)} & \multicolumn{3}{c}{FB15K-DB15K (80\%)} \\
        \cmidrule(r){2-4} \cmidrule(lr){5-7} \cmidrule(l){8-10}
        Methods & MRR(\%) & Hits@1(\%) & Hits@10(\%) & MRR(\%) & Hits@1(\%) & Hits@10(\%) & MRR(\%) & Hits@1(\%) & Hits@10(\%) \\
        \midrule

        EVA & 18.70 & 12.43 & 31.83 & 25.3 & 17.7 & 40.5 & 35.3 & 25.8 & 53.3\\
        MCLEA & 54.0 & 44.89 & 71.54 & 64.9 & 56.11 & 78.87 & 77.9 & 70.97 & 88.17\\
        MSNEA & 60.76 & 52.21 & 70.39 & 64.23 & 56.04 & 79.89 & 63.71 & 55.79 & 78.62\\
        MEAformer & 66.3 & 57.72 & 81.27 & 74.8 & 68.21 & 86.49 & 82.8 & 77.55 & 88.92\\
        SimDiff & 67.8 & 60.16 & 81.89 & 77.1 & 71.55 & 87.21 & 84.41 & 80.79 & 91.01\\
        \midrule
        ALMEA$_{\text{w/ CIS}}$ & 59.43 & 50.79 & 75.64 & 74.16 & 67.40 & 86.16 & 83.80 & 79.08 & \textbf{92.41}\\
        ALMEA$_{\text{w/ IL}}$ & \textbf{69.57} & \textbf{62.83} & \textbf{83.99} & \textbf{77.8} & \textbf{72.76} & \textbf{87.52} & \textbf{84.98} & \textbf{81.01} & 91.70\\

    \end{tabular}}
    
    \resizebox{1\textwidth}{!}{
    \begin{tabular}{lccccccccc}
        \toprule
        & \multicolumn{3}{c}{FB15K-YAGO15K (20\%)} & \multicolumn{3}{c}{FB15K-YAGO15K (50\%)} & \multicolumn{3}{c}{FB15K-YAGO15K (80\%)} \\
        \cmidrule(r){2-4} \cmidrule(lr){5-7} \cmidrule(l){8-10}
        Methods & MRR(\%) & Hits@1(\%) & Hits@10(\%) & MRR(\%) & Hits@1(\%) & Hits@10(\%) & MRR(\%) & Hits@1(\%) & Hits@10(\%) \\
        \midrule
        EVA & 15.3 & 9.9 & 26.15 & 23.8 & 16.39 & 38.45 & 38.5 & 30.4 & 53.9\\
        MCLEA & 50.6 & 42.43 & 66.09 & 63.6 & 56.68 & 76.25 & 71.1 & 64.64 & 83.79\\
        MSNEA & 55.25 & 46.93 & 64.58 & 57.57 & 49.65 & 72.41 & 57.59 & 49.47 & 66.53\\
        MEAformer & 53.6 & 44.9 & 69.85 & 67.9 & 60.43 & 81.43 & 77.69 & 72.1 & 85.36\\
        SimDiff & 58.65 & 52.68 & 71.88 & 71.57 & 64.79 & 82.97 & 78.25 & 73.35 & 86.81\\
        \midrule
        ALMEA$_{\text{w/ CIS}}$ & 50.92 & 42.11 & 67.70 & 68.77 & 61.33 & 82.50 & 78.34 & 72.27 & 88.95\\
        ALMEA$_{\text{w/ IL}}$ & \textbf{59.51} & \textbf{53.58} & \textbf{72.45} & \textbf{71.97} & \textbf{65.19} & \textbf{83.37} & \textbf{79.97} & \textbf{74.46} & \textbf{89.73}\\
        \bottomrule
    \end{tabular}}
\end{table*}
In the current MMKG literature, we note that SimDiff (the current state-of-the-art method), MEAformer, MCLEA, and MSNEA all adopt the Iterative Learning (IL) paradigm. However, we have identified issues with the IL paradigm: all candidate pairs are added to the training set as pseudo-labels. This implies that IL operates as a purely self-training method, unconstrained by annotation budgets and offering no guarantee regarding the quality of the pseudo-labels. This is one of the key reasons for our proposal of Controlled Iterative Selection (with a focus on ensuring the quality of candidate pairs).

All baseline models utilising IL were run under an unconstrained budget setting. To ensure fairness and comparability under equivalent conditions, we reclassified all IL-based baseline models to the unconstrained budget setting. Under this setting, we denote ALMEA with Iterative Learning (using an unlimited budget) as ALMEA$_{\text{w/ IL}}$, and for all baseline models and ALMEA$_{\text{w/ IL}}$, had different budget caps. We found that as the diagonal diversity-weight matrix $D$ continues to constrain the direction of sparsification, ALMEA$_{\text{w/ IL}}$’s budget cap is consistently lower than that of other IL-based baseline models. In other words, the information gain from new candidate pairs that ALMEA$_{\text{w/ IL}}$ can acquire during the iterative process is, in fact, more limited.

However, even under these conditions, ALMEA$_{\text{w/ IL}}$ matches or outperforms all IL-based baseline models, as shown in Table~\ref{CIS under Unlimited Budget}. The fundamental reason is that ALMEA$_{\text{w/ IL}}$ effectively circumvents two issues in the IL process: firstly, the quality of candidate pairs gradually declines over iterations; and secondly, the repeated sampling of duplicate candidate pairs that have already been labelled.

\subsection{Robustness of CIS under Candidate-space Imbalance}\label{Robustness of CIS}
\begin{table}[htbp]
  \centering
  \setlength\tabcolsep{2pt}
  \caption{CIS robustness under imbalanced candidate spaces. We gradually reduce the number of target entities to simulate the scale difference between the source and target KGs, and report the resulting alignment performance.}
  \label{Robustness of CIS under Candidate-space Imbalance}
  \small
    \begin{tabular}{lccc}
        \toprule
        & \multicolumn{3}{c}{FB15K-DB15K (20\%)} \\
        \cmidrule(r){2-4} 
        Methods & MRR(\%) & Hits@1(\%) & Hits@10(\%) \\
        \midrule
        
        ALMEA (Imbalance-0.05$\times$) & 55.06 & 45.73 & 72.80 \\
        ALMEA (Imbalance-0.1$\times$) & 55.51 & 46.21 & 73.36 \\
        ALMEA (Imbalance-0.5$\times$) & 59.09 & 50.27 & 75.76 \\
        ALMEA & \textbf{59.43} & \textbf{50.79} & \textbf{75.64} \\
        \bottomrule
    \end{tabular}
\end{table}
Although CIS is designed to mitigate \textbf{suppression} under limited supervision, it fundamentally relies on the availability of candidate pairs within the candidate space. If entities from the source or target knowledge graphs are unavailable in a candidate pair, this directly affects the quality of candidate pairs selected by CIS. However, in real-world MMEA scenarios, the sizes of the source and target knowledge graphs are often highly imbalanced. For example, As shown in Table \ref{Statist_datasets} the FB15K-DB15K dataset, the total number of entities on the Source set is 14,951, while the Target set contains 12,846 entities. It is therefore necessary to examine whether CIS remains reliable when the availability of entities in the target kg continues to deteriorate. To simulate the disparity in scale between Source and Target, we reduced the total number of Target entities by factors of 0.5, 0.1, and 0.05, respectively, relative to the total on the Source side. These settings are denoted as ALMEA (Imbalance-0.5$\times$), ALMEA (Imbalance-0.1$\times$), and ALMEA (Imbalance-0.05$\times$), respectively.

As shown in Table \ref{Robustness of CIS under Candidate-space Imbalance}, we observe minimal performance degradation when data imbalance is set to 0.5$\times$. After randomly masking half of the Target entities, CIS still finds it easier to predict matching entities from the Source KG to the Target KG. Most ground-truth entities remain discoverable within the candidate space, as CIS filters out noise while preserving multimodal semantic information. When the Target total was reduced to 0.1$\times$, most Source KG entities could no longer find matching counterparts in the Target KG. Constructing the dissimilarity matrix $A$ became divergent and ambiguous, resulting in a significant decline in alignment performance. The same applies at 0.05$\times$.

\clearpage
\section{Notation Summary}\label{sec:notation_summary}
Table~\ref{tab:MMKG notation} summarizes the notation used for the MMKG setting, modality embeddings, the CMI and LSC components, CIS optimization, and evaluation metrics.
\begingroup
\small
\renewcommand{\arraystretch}{1.05}
\setlength{\LTleft}{\fill}
\setlength{\LTright}{\fill}
\begin{longtable}{@{}>{\centering\arraybackslash}p{0.18\textwidth}p{0.76\textwidth}@{}}
  \caption{Summary of Notations Used in ALMEA}
  \label{tab:MMKG notation}
  \label{tab:MEC notation}
  \label{tab:LSL notation}
  \label{tab:LSC notation}
  \label{tab:ACS notation}
  \label{tab:EMN notation}\\
  \toprule
  \textbf{Notation} & \textbf{Explanation} \\
  \midrule
  \endfirsthead
  \multicolumn{2}{c}{\tablename~\thetable\ (continued)} \\
  \toprule
  \textbf{Notation} & \textbf{Explanation} \\
  \midrule
  \endhead
  \midrule
  \multicolumn{2}{r}{Continued on next page} \\
  \endfoot
  \bottomrule
  \endlastfoot

  \multicolumn{2}{@{}l}{\textbf{Multimodal Knowledge Graph (MMKG)}} \\*
  $\mathcal{G}$ & Multimodal knowledge graph \\
  $\mathcal{E}$ & Set of entities \\
  $\mathcal{R}$ & Set of relations \\
  $\mathcal{A}$ & Set of attributes \\
  $\mathcal{V}$ & Visual information associated with the entities \\
  $\mathcal{T}$ & Set of relational triples \\
  $\mathcal{G}_S, \mathcal{G}_T$ & Source and target MMKGs \\

  \midrule
  \multicolumn{2}{@{}l}{\textbf{Input and Modality Embeddings}} \\*
  $x^g$ & Original neighborhood feature vector \\
  $x^r$ & Original relation feature vector \\
  $x^a$ & Original attribute feature vector \\
  $x^v$ & Original visual feature vector \\
  $h^g$ & Neighborhood modality embedding \\
  $h^r$ & Relation modality embedding \\
  $h^a$ & Attribute modality embedding \\
  $h^v$ & Visual modality embedding \\
  $W_g$ & Neighborhood linear-transformation weights \\
  $W_r$ & Relation linear-transformation weights \\
  $W_a$ & Attribute linear-transformation weights \\
  $W_v$ & Visual linear-transformation weights \\
  $\mathcal{M}$ & Pretrained multimodal encoder \\
  $\mathrm{GAT}$ & Graph attention network transformation \\
  $\mathrm{Concat}$ & Concatenation operation \\

  \midrule
  \multicolumn{2}{@{}l}{\textbf{Cross-Modality Imputation (CMI)}} \\*
  $\tilde{h}^g$ & Masked neighborhood modality embedding \\
  $\tilde{h}^r$ & Masked relation modality embedding \\
  $\tilde{h}^a$ & Masked attribute modality embedding \\
  $\tilde{h}^v$ & Masked visual modality embedding \\
  $\mathbf{h}^g$ & Hidden neighborhood representation \\
  $\mathbf{h}^r$ & Hidden relation representation \\
  $\mathbf{h}^a$ & Hidden attribute representation \\
  $\mathbf{h}^v$ & Hidden visual representation \\
  $z^g$ & Latent neighborhood representation \\
  $z^r$ & Latent relation representation \\
  $z^a$ & Latent attribute representation \\
  $z^v$ & Latent visual representation \\
  $\mathbf{z}^g$ & Calibrated neighborhood latent embedding \\
  $\mathbf{z}^r$ & Calibrated relation latent embedding \\
  $\mathbf{z}^a$ & Calibrated attribute latent embedding \\
  $\mathbf{z}^v$ & Calibrated visual latent embedding \\
  $\mathcal{N}(\mu,\sigma^2)$ & Gaussian distribution with mean $\mu$ and variance $\sigma^2$ \\
  $\epsilon\sim\mathcal{N}(0,1)$ & Standard normal noise used for reparameterization \\
  $\mathcal{L}_{DML}$ & Distribution matching loss \\
  $\mathcal{L}_{URL}$ & Unimodal reconstruction loss \\
  $\mathrm{Encoder}$ & Variational autoencoder encoder \\
  $\mathrm{Decoder}$ & Variational autoencoder decoder \\

  \midrule
  \multicolumn{2}{@{}l}{\textbf{Latent Semantic Calibration (LSC)}} \\*
  $\mathrm{Tanh}$ & Hyperbolic tangent activation function \\
  $\mathrm{MLP}$ & Multilayer perceptron \\
  $\mathrm{MSE}$ & Mean squared error \\
  $\mathcal{L}_{SCL}$ & Semantic calibration loss \\
  $\mathcal{L}_{CMAL}$ & Joint alignment loss \\
  $\lambda_1,\lambda_2,\lambda_3,\lambda_4$ & Weights for SCL, DML, URL, and CMAL, respectively \\

  \midrule
  \multicolumn{2}{@{}l}{\textbf{Controlled Iterative Selection (CIS)}} \\*
  $L$ & Current set of labeled entity pairs \\
  $Q$ & Candidate pool of potential entity pairs \\
  $Q_{\text{sorted}}$ & Candidate pool sorted by confidence \\
  $K=|Q_{\text{sorted}}|$ & Number of candidate pairs \\
  $Z \in \mathbb{R}^{K \times K}$ & Sparse representation matrix \\
  $A \in \mathbb{R}^{K \times K}$ & Pairwise candidate dissimilarity matrix \\
  $C \in \mathbb{R}^{K \times |L|}$ & Distances between candidate pairs and labeled pairs \\
  $\mathbf{d} \in \mathbb{R}^{K}$ & Vector of candidate-level diversity weights \\
  $D=\operatorname{diag}(\mathbf{d}) \in \mathbb{R}^{K \times K}$ & Diagonal diversity-weight matrix \\
  $\alpha$ & Sparsification factor \\
  $\tau$ & Confidence threshold \\
  $P \in \mathbb{R}^{K \times K}$ & Auxiliary copy of $Z$ in ADMM optimization \\
  $T_{\mathrm{iter}}$ & Maximum number of ADMM iterations \\
  $\mu \in \mathbb{R}^{K \times K}$ & Dual variable associated with the constraint $Z=P$ \\
  $\rho$ & ADMM penalty parameter \\
  $\|\cdot\|_{2,1}$ & Mixed $\ell_{2,1}$ norm used for row sparsity \\
  $\|\cdot\|_F^2$ & Squared Frobenius norm \\
  $\operatorname{tr}(\cdot)$ & Matrix trace \\

  \midrule
  \multicolumn{2}{@{}l}{\textbf{Evaluation Metrics}} \\*
  $\text{MRR}$ & Mean Reciprocal Rank \\
  $\text{Hits@1}$ & Hits at Rank 1 \\
  $\text{Hits@10}$ & Hits at Rank 10 \\
  $\text{TPR}$ & True Positive Rate \\
  $\text{FPR}$ & False Positive Rate \\
\end{longtable}
\endgroup

%% file: main.bbl
\begin{thebibliography}{10}
\providecommand{\url}[1]{#1}
\csname url@samestyle\endcsname
\providecommand{\newblock}{\relax}
\providecommand{\bibinfo}[2]{#2}
\providecommand{\BIBentrySTDinterwordspacing}{\spaceskip=0pt\relax}
\providecommand{\BIBentryALTinterwordstretchfactor}{4}
\providecommand{\BIBentryALTinterwordspacing}{\spaceskip=\fontdimen2\font plus
\BIBentryALTinterwordstretchfactor\fontdimen3\font minus
  \fontdimen4\font\relax}
\providecommand{\BIBforeignlanguage}[2]{{%
\expandafter\ifx\csname l@#1\endcsname\relax
\typeout{** WARNING: IEEEtran.bst: No hyphenation pattern has been}%
\typeout{** loaded for the language `#1'. Using the pattern for}%
\typeout{** the default language instead.}%
\else
\language=\csname l@#1\endcsname
\fi
#2}}
\providecommand{\BIBdecl}{\relax}
\BIBdecl

\bibitem{chen2020mmea}
L.~Chen, Z.~Li, Y.~Wang, T.~Xu, Z.~Wang, and E.~Chen, ``Mmea: entity alignment
  for multi-modal knowledge graph,'' in \emph{Knowledge Science, Engineering
  and Management: 13th International Conference, KSEM 2020, Hangzhou, China,
  August 28--30, 2020, Proceedings, Part I 13}.\hskip 1em plus 0.5em minus
  0.4em\relax Springer, 2020, pp. 134--147.

\bibitem{liu2021visual}
F.~Liu, M.~Chen, D.~Roth, and N.~Collier, ``Visual pivoting for (unsupervised)
  entity alignment,'' in \emph{Proceedings of the AAAI conference on artificial
  intelligence}, vol.~35, no.~5, 2021, pp. 4257--4266.

\bibitem{chen2022multi}
L.~Chen, Z.~Li, T.~Xu, H.~Wu, Z.~Wang, N.~J. Yuan, and E.~Chen, ``Multi-modal
  siamese network for entity alignment,'' in \emph{Proceedings of the 28th ACM
  SIGKDD conference on knowledge discovery and data mining}, 2022, pp.
  118--126.

\bibitem{lin2022multi}
Z.~Lin, Z.~Zhang, M.~Wang, Y.~Shi, X.~Wu, and Y.~Zheng, ``Multi-modal
  contrastive representation learning for entity alignment,'' \emph{arXiv
  preprint arXiv:2209.00891}, 2022.

\bibitem{chen2023meaformer}
Z.~Chen, J.~Chen, W.~Zhang, L.~Guo, Y.~Fang, Y.~Huang, Y.~Zhang, Y.~Geng, J.~Z.
  Pan, W.~Song \emph{et~al.}, ``Meaformer: Multi-modal entity alignment
  transformer for meta modality hybrid,'' in \emph{Proceedings of the 31st ACM
  International Conference on Multimedia}, 2023, pp. 3317--3327.

\bibitem{guo2023revisit}
L.~Guo, Z.~Chen, J.~Chen, Y.~Fang, W.~Zhang, and H.~Chen, ``Revisit and
  outstrip entity alignment: A perspective of generative models,'' \emph{arXiv
  preprint arXiv:2305.14651}, 2023.

\bibitem{chen2023rethinking}
Z.~Chen, L.~Guo, Y.~Fang, Y.~Zhang, J.~Chen, J.~Z. Pan, Y.~Li, H.~Chen, and
  W.~Zhang, ``Rethinking uncertainly missing and ambiguous visual modality in
  multi-modal entity alignment,'' in \emph{International Semantic Web
  Conference}.\hskip 1em plus 0.5em minus 0.4em\relax Springer, 2023, pp.
  121--139.

\bibitem{sun2018bootstrapping}
Z.~Sun, W.~Hu, Q.~Zhang, and Y.~Qu, ``Bootstrapping entity alignment with
  knowledge graph embedding.'' in \emph{Ijcai}, vol.~18, no. 2018, 2018.

\bibitem{kipf2017semi}
\BIBentryALTinterwordspacing
T.~N. Kipf and M.~Welling, ``Semi-supervised classification with graph
  convolutional networks,'' in \emph{Proceedings of the International
  Conference on Learning Representations (ICLR)}, 2017. [Online]. Available:
  \url{https://arxiv.org/abs/1609.02907}
\BIBentrySTDinterwordspacing

\bibitem{velickovic2018graph}
\BIBentryALTinterwordspacing
P.~Veli{\v{c}}kovi{\'c}, G.~Cucurull, A.~Casanova, A.~Romero, P.~Li{\`o}, and
  Y.~Bengio, ``Graph attention networks,'' in \emph{Proceedings of the
  International Conference on Learning Representations (ICLR)}, 2018. [Online].
  Available: \url{https://arxiv.org/abs/1710.10903}
\BIBentrySTDinterwordspacing

\bibitem{bordes2013translating}
A.~Bordes, N.~Usunier, A.~Garcia-Duran, J.~Weston, and O.~Yakhnenko,
  ``Translating embeddings for modeling multi-relational data,'' \emph{Advances
  in neural information processing systems}, vol.~26, 2013.

\bibitem{wang2014knowledge}
Z.~Wang, J.~Zhang, J.~Feng, and Z.~Chen, ``Knowledge graph embedding by
  translating on hyperplanes,'' in \emph{Proceedings of the AAAI conference on
  artificial intelligence}, vol.~28, no.~1, 2014.

\bibitem{li2023multi}
Q.~Li, C.~Ji, S.~Guo, Z.~Liang, L.~Wang, and J.~Li, ``Multi-modal knowledge
  graph transformer framework for multi-modal entity alignment,'' \emph{arXiv
  preprint arXiv:2310.06365}, 2023.

\bibitem{ni2023psnea}
W.~Ni, Q.~Xu, Y.~Jiang, Z.~Cao, X.~Cao, and Q.~Huang, ``Psnea: Pseudo-siamese
  network for entity alignment between multi-modal knowledge graphs,'' in
  \emph{Proceedings of the 31st ACM International Conference on Multimedia},
  2023, pp. 3489--3497.

\bibitem{li2024simdiff}
R.~Li, S.~Di, L.~Chen, and X.~Zhou, ``Simdiff: Simple denoising probabilistic
  latent diffusion model for data augmentation on multi-modal knowledge
  graph,'' in \emph{Proceedings of the 30th ACM SIGKDD Conference on Knowledge
  Discovery and Data Mining}, 2024, pp. 1631--1642.

\bibitem{wang2025otmea}
C.~Wang, W.~Wang, X.~Li, Q.~Liang, and F.~Bao, ``Otmea: Multi-modal entity
  alignment via optimal transport,'' in \emph{ICASSP 2025-2025 IEEE
  International Conference on Acoustics, Speech and Signal Processing
  (ICASSP)}.\hskip 1em plus 0.5em minus 0.4em\relax IEEE, 2025, pp. 1--5.

\bibitem{li2023attribute}
Q.~Li, S.~Guo, Y.~Luo, C.~Ji, L.~Wang, J.~Sheng, and J.~Li,
  ``Attribute-consistent knowledge graph representation learning for
  multi-modal entity alignment,'' in \emph{Proceedings of the ACM Web
  Conference 2023}, 2023, pp. 2499--2508.

\bibitem{chen2020uniter}
Y.-C. Chen, L.~Li, L.~Yu, A.~El~Kholy, F.~Ahmed, Z.~Gan, Y.~Cheng, and J.~Liu,
  ``Uniter: Universal image-text representation learning,'' in \emph{European
  conference on computer vision}.\hskip 1em plus 0.5em minus 0.4em\relax
  Springer, 2020, pp. 104--120.

\bibitem{li2021align}
J.~Li, R.~Selvaraju, A.~Gotmare, S.~Joty, C.~Xiong, and S.~C.~H. Hoi, ``Align
  before fuse: Vision and language representation learning with momentum
  distillation,'' \emph{Advances in neural information processing systems},
  vol.~34, pp. 9694--9705, 2021.

\bibitem{cohen1997align}
G.~H. Cohen, ``Align: a program to superimpose protein coordinates, accounting
  for insertions and deletions,'' \emph{Applied Crystallography}, vol.~30,
  no.~6, pp. 1160--1161, 1997.

\bibitem{berrendorf2021active}
M.~Berrendorf, E.~Faerman, and V.~Tresp, ``Active learning for entity
  alignment,'' in \emph{Advances in Information Retrieval: 43rd European
  Conference on IR Research, ECIR 2021, Virtual Event, March 28--April 1, 2021,
  Proceedings, Part I 43}.\hskip 1em plus 0.5em minus 0.4em\relax Springer,
  2021, pp. 48--62.

\bibitem{liu2021activeea}
B.~Liu, H.~Scells, G.~Zuccon, W.~Hua, and G.~Zhao, ``Activeea: Active learning
  for neural entity alignment,'' \emph{arXiv preprint arXiv:2110.06474}, 2021.

\bibitem{zeng2021reinforced}
W.~Zeng, X.~Zhao, J.~Tang, and C.~Fan, ``Reinforced active entity alignment,''
  in \emph{Proceedings of the 30th ACM international conference on information
  \& knowledge management}, 2021, pp. 2477--2486.

\bibitem{huang2023deep}
J.~Huang, Z.~Sun, Q.~Chen, X.~Xu, W.~Ren, and W.~Hu, ``Deep active alignment of
  knowledge graph entities and schemata,'' \emph{Proceedings of the ACM on
  Management of Data}, vol.~1, no.~2, pp. 1--26, 2023.

\bibitem{wu2024deep}
R.~Wu, H.~Wang, H.-T. Chen, and G.~Carneiro, ``Deep multimodal learning with
  missing modality: A survey,'' \emph{arXiv preprint arXiv:2409.07825}, 2024.

\bibitem{Garcia-DuranN18}
A.~Garc{\'{\i}}a{-}Dur{\'{a}}n and M.~Niepert, ``Kblrn: End-to-end learning of
  knowledge base representations with latent, relational, and numerical
  features,'' in \emph{Proceedings of the Thirty-Fourth Conference on
  Uncertainty in Artificial Intelligence, {UAI}}.\hskip 1em plus 0.5em minus
  0.4em\relax {AUAI} Press, 2018, pp. 372--381.

\bibitem{yuan2010multiple}
Y.~C. Yuan, ``Multiple imputation for missing data: Concepts and new
  development (version 9.0),'' \emph{SAS Institute Inc, Rockville, MD},
  vol.~49, no. 1-11, p.~12, 2010.

\bibitem{vaswani2017attention}
A.~Vaswani, ``Attention is all you need,'' \emph{Advances in Neural Information
  Processing Systems}, 2017.

\bibitem{gong2025multimodal}
B.~Gong, X.~Gao, P.~Zhu, Q.~Hu, and B.~Cao, ``Multimodal negative learning,''
  \emph{arXiv preprint arXiv:2510.20877}, 2025.

\bibitem{kingma2013auto}
D.~P. Kingma, ``Auto-encoding variational bayes,'' \emph{arXiv preprint
  arXiv:1312.6114}, 2013.

\bibitem{elhamifar2015dissimilarity}
E.~Elhamifar, G.~Sapiro, and S.~S. Sastry, ``Dissimilarity-based sparse subset
  selection,'' \emph{IEEE transactions on pattern analysis and machine
  intelligence}, vol.~38, no.~11, pp. 2182--2197, 2015.

\bibitem{sun2017cross}
Z.~Sun, W.~Hu, and C.~Li, ``Cross-lingual entity alignment via joint
  attribute-preserving embedding,'' in \emph{International semantic web
  conference}.\hskip 1em plus 0.5em minus 0.4em\relax Springer, 2017, pp.
  628--644.

\bibitem{wang2018cross}
Z.~Wang, Q.~Lv, X.~Lan, and Y.~Zhang, ``Cross-lingual knowledge graph alignment
  via graph convolutional networks,'' in \emph{Proceedings of the 2018
  conference on empirical methods in natural language processing}, 2018, pp.
  349--357.

\bibitem{pei2019semi}
S.~Pei, L.~Yu, R.~Hoehndorf, and X.~Zhang, ``Semi-supervised entity alignment
  via knowledge graph embedding with awareness of degree difference,'' in
  \emph{The world wide web conference}, 2019, pp. 3130--3136.

\bibitem{student1908probable}
Student, ``The probable error of a mean,'' \emph{Biometrika}, pp. 1--25, 1908.

\bibitem{demvsar2006statistical}
J.~Dem{\v{s}}ar, ``Statistical comparisons of classifiers over multiple data
  sets,'' \emph{Journal of Machine learning research}, vol.~7, no. Jan, pp.
  1--30, 2006.

\bibitem{OoroRubio2019AnsweringVQ}
D.~O{\~n}oro-Rubio, M.~Niepert, A.~Garc{\'i}a-Dur{\'a}n, R.~Gonzalez-Sanchez,
  and R.~J. L{\'o}pez-Sastre, ``Answering visual-relational queries in
  web-extracted knowledge graphs,'' in \emph{AKBC}, 2019.

\bibitem{liu2019mmkg}
Y.~Liu, H.~Li, A.~Garcia-Duran, M.~Niepert, D.~Onoro-Rubio, and D.~S.
  Rosenblum, ``Mmkg: multi-modal knowledge graphs,'' in \emph{European Semantic
  Web Conference}.\hskip 1em plus 0.5em minus 0.4em\relax Springer, 2019, pp.
  459--474.

\bibitem{chen2020simple}
T.~Chen, S.~Kornblith, M.~Norouzi, and G.~Hinton, ``A simple framework for
  contrastive learning of visual representations,'' in \emph{International
  conference on machine learning}.\hskip 1em plus 0.5em minus 0.4em\relax PMLR,
  2020, pp. 1597--1607.

\bibitem{cao2022otkge}
Z.~Cao, Q.~Xu, Z.~Yang, Y.~He, X.~Cao, and Q.~Huang, ``Otkge: Multi-modal
  knowledge graph embeddings via optimal transport,'' \emph{Advances in neural
  information processing systems}, vol.~35, pp. 39\,090--39\,102, 2022.

\bibitem{settles2009active}
B.~Settles, ``Active learning literature survey,'' 2009.

\bibitem{phoenix1992elements}
S.~Phoenix, ``Elements of information theory,'' 1992.

\end{thebibliography}
